\documentclass[]{article}
\usepackage{xurl}

\newif\ifispublish 
\ispublishtrue

\usepackage[subtle,bibliography=normal]{savetrees}

\usepackage{microtype}
\usepackage{graphicx}
\setkeys{Gin}{draft=false}
\usepackage{amsmath}
\usepackage{amssymb}
\usepackage{dsfont}     
\usepackage[table]{xcolor}   
\usepackage{booktabs}
\usepackage{colortbl}
\usepackage{arydshln}        
\usepackage{multirow}
\usepackage{placeins}       
\usepackage{hyperref}      
\usepackage{cleveref}
\crefname{ALC@unique}{line}{lines}
\Crefname{ALC@unique}{Line}{Lines}
\crefname{appendix}{appendix}{appendices}
\Crefname{appendix}{Appendix}{Appendices}

\crefformat{section}{§#2#1#3}
\crefformat{subsection}{§#2#1#3}
\crefformat{subsubsection}{§#2#1#3}

\crefmultiformat{section}{§§#2#1#3}{ and~#2#1#3}{, #2#1#3}{ and~#2#1#3}
\crefrangemultiformat{section}{§§#2#1#3}{ and~#2#1#3}{, #2#1#3}{ and~#2#1#3}
\crefrangeformat{section}{§§#2#1#3\crefrangeconjunction#4#2#3}

\usepackage{xspace}

\graphicspath{{Figures/}}

\ifispublish
    \usepackage[disable]{todonotes}			
\else
    \usepackage[]{todonotes}
\fi

\usepackage[preprint]{mlsys2025}
\hypersetup{pdfsubject={Preprint}}
\makeatletter
\renewcommand{\theALC@unique}{\arabic{ALC@line}}
\renewcommand{\theHALC@unique}{\arabic{ALC@unique}}
\makeatother

\usepackage{pifont}
\newif\ifinternalnotes
\internalnotestrue

\newcommand{\para}[1]{\vspace{-1pt}\noindent\textbf{\textit{#1}.}}

\newcommand{\eg}{e.g.,\xspace}

\newcommand{\ignore}[1]{}

\ifispublish
    
    \newcommand{\TODO}[1]{}
    \newcommand{\assign}[2]{}

\else
    \newcommand{\TODO}[1]{{\textcolor{red}{[\textbf{TODO:} #1]}}}

    \newcommand{\assign}[2]{{\colorbox{cyan}{[for \textbf{#1:} #2]}}}
\fi

\def\addauthnote#1#2{%
	\expandafter\def\csname#1\endcsname##1{%
	\todo[size=\footnotesize,color=#2,inline]
			{\textbf{\underline{\texttt{#1}}:} ##1}\xspace}
   \expandafter\def\csname#1li\endcsname##1{%
    \todo[size=\footnotesize,color=#2,inline,inlinewidth=5cm, noinlinepar]
			{\textbf{\underline{\texttt{#1}}:} ##1}\xspace}
    \expandafter\def\csname#1Res\endcsname##1{%
	\todo[size=\footnotesize,color=gray,inline]
			{\textbf{\underline{[Resolved] \texttt{#1}}:} ##1}\xspace}
}

\newcommand{\Stat}[1]{#1\xspace}

\DeclareRobustCommand{\circled}[1]{%
  {\scriptsize
   \tikz[baseline=(char.base)]{
     \node[shape=circle,draw,scale=1.2,inner sep=1pt] (char) {#1};
   }}%
}

\ignore{
\newcommand{\mintedfsize}{\footnotesize}
\newmintinline[inlPy]{python}{fontsize=\mintedfsize, escapeinside=||}
\newmintinline[inlJava]{java}{fontsize=\mintedfsize, escapeinside=||}
\newmintinline[inlC]{C}{fontsize=\mintedfsize, escapeinside=||}
\newmintinline[inlPhp]{php}{fontsize=\mintedfsize, escapeinside=||}
}

\def\shac#1{
	\expandafter\def\csname#1\endcsname{\ac{#1}\xspace}
    \expandafter\def\csname#1c\endcsname{\Ac{#1}\xspace}{}
    \expandafter\def\csname#1s\endcsname{\acp{#1}\xspace}{}
}
\newcommand{\ind}{\mathds{1}}
\newcommand{\Acc}{\mathrm{Acc}}
\newcommand{\LAST}[2]{\mathrm{LAST}_{#1,#2}}
\newcommand{\ANY}[2]{\mathrm{ANY}_{#1,#2}}
\newcommand{\CNT}[2]{\mathrm{COUNT2}_{#1,#2}}
\newcommand{\bitprobe}{\textsc{GateWatch}\xspace}
\newcommand{\bitprobes}{\textsc{GateWatch} probes\xspace}
\newcommand{\bitprobeserv}{\textsc{GateWatchServe}\xspace}
\newcommand{\bitprobeserve}{\textsc{GateWatchServe}\xspace}
\newcommand{\bitprobebuilder}{\textsc{GateWatchBuilder}\xspace}

\newcommand{\bitprobebuild}{\textsc{GateWatchBuilder}\xspace}
\newcommand{\MultiTokenFeatureShare}{\Stat{61\%}}

\definecolor{win}{RGB}{214,239,214}     
\definecolor{ours}{RGB}{253,236,221}    
\definecolor{grouphead}{RGB}{242,242,242}
\newcommand{\dashrule}{\noalign{\vskip 1.5pt}\hdashline\noalign{\vskip 2pt}}
\newcommand{\grp}[2]{\multicolumn{#1}{@{}l}{\cellcolor{grouphead}\textit{#2}}}
\makeatletter
\newcommand{\teasercaption}[1]{\def\@captype{figure}\caption{#1}}
\makeatother

\newcommand{\sys}{\bitprobe}
\def\ptitle{GateWatch: Low-Overhead LLM Monitoring with Boolean Circuits}
\def\pkeywords{}

\mlsystitlerunning{GateWatch: Low-Overhead LLM Monitoring}

\hypersetup{
	colorlinks	= true,
	citecolor	= {red!70!black},
	filecolor	= black,
	linkcolor	= {green!70!black},
	urlcolor	= {blue!70!black},
	pagebackref	= false,
	pdftitle	= {\ptitle},
	pdfkeywords	= {\pkeywords}
}

\addauthnote{firas}{blue!25}
\addauthnote{yaniv}{green!25}
\addauthnote{TODO}{yellow!25}

\ifispublish
    \newcommand{\draftunboundedpages}{}
\else
    \newcommand{\draftunboundedpages}{%
        \fancyfoot[C]{\footnotesize\thepage/\ensuremath{\infty}}%
    }
\fi

\begin{document}

\twocolumn[

\mlsystitle{\ptitle}

\mlsyssetsymbol{equal}{*}
\begin{mlsysauthorlist}
\mlsysauthor{Firas Spanioli}{Technion}
\mlsysauthor{Yaniv David}{Technion}
\end{mlsysauthorlist}
\mlsysaffiliation{Technion}{Technion - Israel Institute of Technology}
\mlsyscorrespondingauthor{Firas Spanioli}{sfirasss@campus.technion.ac.il}
\mlsyscorrespondingauthor{Yaniv David}{yanivmd@technion.ac.il}
\mlsyskeywords{Machine Learning, Sparse Autoencoders, Probes, Logic Synthesis}

\vskip 0.3in


\begin{abstract}
As large language models (LLMs) and LLM-based agents are increasingly deployed, monitoring their behavior becomes important for detecting unsafe or unintended actions. Activation probes offer an efficient approach, but existing methods face trade-offs between accuracy, interpretability, and computational overhead. Probes on raw activations lack interpretable inputs, whereas sparse autoencoder (SAE) probes use interpretable features but often rely on additive classifiers and large SAE encoders, limiting interpretability and increasing memory overhead.
We introduce \bitprobe{}, a compact, interpretable classifier that expresses its entire decision rule as an explicit Boolean logic over a small number of binarized SAE features.
We explore three approaches to constructing these circuits under a specified gate budget. For efficient deployment, \bitprobeserve{} computes only selected SAE features, enables streaming of their binary values across layers during inference, and evaluates multiple \bitprobe{} classifiers in parallel using compact lookup tables.
Across 24 tasks on Gemma-2-9B, \bitprobe{} achieves 80.49\% mean test accuracy using only 3.8 SAE features and 3.5 logic gates per probe on average, just 1.17 percentage points below full 16K-feature SAE probes, while matching or outperforming them on 10 tasks and reducing probe memory by 99.98\%. Inside vLLM, 1{,}000 concurrent multi-layer, multi-token \bitprobe{} classifiers retain 97--98\% of baseline throughput and 99.8\% of KV-cache capacity. 
\end{abstract}

\vskip 0.1in
]

\printAffiliationsAndNotice{}

\section{Introduction}
\label{sec:intro}

\begin{figure}[t]
\centering
\includegraphics[width=\columnwidth]{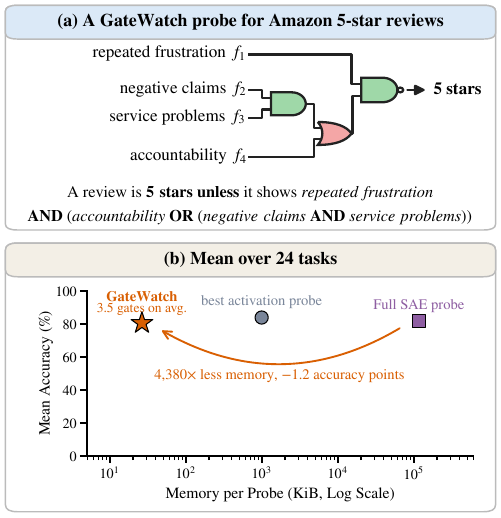}
\caption{\textbf{\sys at a glance.} (a)~A complete \bitprobe: 
it reaches 95.49\% accuracy, against 89.44\% for the 16K-SAE probe. 
(b)~ The best activation probe is more accurate but has no readable rule.}

\label{fig:teaser}
\end{figure}

Probes can help detect unsafe, unreliable, or unintended behavior in LLMs.
As LLMs are increasingly incorporated in real-world applications or deployed as
LLM-based agents, monitoring them becomes paramount.
This is especially true for LLMs that gain access to sensitive data and
organizational systems, where they may take unauthorized actions, leak data, or
misuse permissions. Activation probes offer one way to detect such behavior:
they are classifiers trained on a model's internal activations
\citep{alain2017probes}.
Because these activations are already computed during inference, probes can be
significantly cheaper than using an external LLM to evaluate the model's
behavior, \eg LLamaGuard \cite{inan2023llamaguard}. Recent work from
Google and Anthropic has explored deploying probes for misuse and jailbreak
detection in user-facing systems
\citep{kramar2026production,cunningham2026constitutional}.

Production monitoring may require many probes to run alongside an LLM, so each
must be accurate enough for reliable decisions, compact enough to coexist with
other monitors, fast enough to preserve serving latency, and interpretable
enough to explain and debug its predictions. Useful signals may appear before
the final token, while averaging over tokens can wash them out
\citep{kramar2026production}. Reading more tokens and layers can increase
feature-extraction work and per-request state.

Existing approaches do not meet these requirements together. Linear and MLP
probes can be accurate, but their dense activation features make decisions
difficult to interpret, and their full-width weights raise the cost of
deploying many monitors. Sparse
autoencoders (SAEs) represent activations with a small number of active
features that can often be given human-readable descriptions
\citep{bricken2023monosemanticity,templeton2024scaling,gao2024scaling}. SAEs can
be used to construct interpretable probes
\citep{gallifant2025saeclassification,shi2025safer}, yet they
do not consistently outperform probes trained directly on model activations,
despite the added cost of computing SAE features
\citep{kantamneni2025saeprobes}.

Moreover, SAE-based probes often rely on linear or additive classifiers to
combine features, which can make decisions hard to interpret and restrict the
interactions they can express. Although these classifiers identify which
features contribute to a prediction, decisions involving many features can
still be difficult to interpret. Additive classifiers also cannot express some
interactions, such as a decision that depends on exactly one of two concepts
being present.

To address these challenges, we introduce \sys, a system for building and 
deploying interpretable probes over selected SAE
features, which we call \bitprobes. As \cref{fig:teaser}a shows, a \bitprobe
represents its entire classification rule as a small Boolean circuit over
binarized SAE features. Each feature becomes a firing bit, and the circuit
combines these bits with simple logic gates (\textsc{and}, \textsc{or},
\textsc{not}, and \textsc{xor}). The explicit rule captures nonlinear feature
interactions while remaining small enough to deploy many probes together.

\bitprobebuild{} selects a small set of SAE features and offers three
alternatives for constructing \bitprobes. \textsc{Compile} starts with a trained
logistic regression classifier over binarized SAE features, converts its
rounded decision rule into an equivalent Boolean circuit using hardware logic
synthesis, and iteratively prunes the circuit based on validation performance
to meet a size budget.
\textsc{Search} uses symbolic regression, while
\textsc{Train} uses differentiable logic gate networks.

\bitprobeserve{} runs the selected rules alongside the LLM in vLLM
\citep{kwon2023vllm}. It computes only the SAE features used by the probes,
tracks their firing across tokens as each selected layer runs, and evaluates
many \bitprobe{} classifiers in parallel as small lookup tables. This path
enable probes to use earlier signals without storing full activation sequences
or computing complete SAE feature vectors during inference.

\Cref{fig:teaser}b summarizes our evaluation of \sys{} on 24 tasks spanning distinct concept families, with
Gemma-2-9B and pretrained
Gemma Scope SAEs, building \bitprobes{} over 3.8 SAE features and 3.5 logic
gates per probe on average that
reach 80.49\% mean test accuracy, 1.17 points below the full 16K-SAE probe
(81.66\%),
match or beat it on 10 tasks, and cut probe memory by 99.98\%.
This compares deployable systems, since with the baselines' own access (layer 20, last token), five-gate Search \bitprobes{} trail the SAE probe by 7.64 points, and reading all layers and tokens recovers most of this gap.
As \cref{sec:ablations} shows, \MultiTokenFeatureShare of selected features
use sequence-wide aggregation rules.
Offering all three construction methods enables task-specific selection:
\textsc{Search}, \textsc{Compile}, and \textsc{Train} produce the most accurate
compact classifiers on 12, 8, and 4 of our 24 tasks, respectively.

To evaluate whether \sys{} can support many concurrent
monitors without slowing down LLM inference,
we stress-test it with 1{,}000 multi-layer, multi-token
\bitprobes.
\sys retains 97--98\% of baseline throughput and
99.8\% of KV-cache capacity.
The interpretable rules also make it easier to understand what drives each
prediction, revealing dataset shortcuts that may be difficult to identify with
dense probes
(\cref{sec:interpretability-results}).

To summarize, this paper makes the following contributions:
\begin{itemize}
\setlength{\itemsep}{0pt}
\setlength{\parsep}{0pt}
\setlength{\topsep}{2pt}
\setlength{\partopsep}{0pt}
\item \textbf{Symbolic Boolean probes.} We introduce \bitprobe{}, a classifier
    whose entire decision rule is represented by an explicit, human-readable
    Boolean logic over multi-layer, multi-token binary SAE features.

\item \textbf{Classifier-to-circuit conversion.} In \bitprobebuilder{}, we develop a pipeline that converts a trained classifier into a Boolean circuit
    through hardware logic synthesis and validation-guided pruning, producing
    circuits under a specified gate budget.

\item \textbf{Implementation and evaluation.} We implement \bitprobeserve on top of vLLM, 
    enabling  streaming of selected SAE features across layers and
    parallel execution of compact \bitprobe{} classifiers. Across 24 tasks, we
    assess accuracy, gate count, memory, serving cost, and the readability of
    complete decision rules.

\end{itemize}


\section{Background}
\label{sec:background}


\subsection{Probes and Sparse Autoencoders}
\label{sec:bg-probes}

\begin{figure*}[t]
\centering
\includegraphics[width=\textwidth]{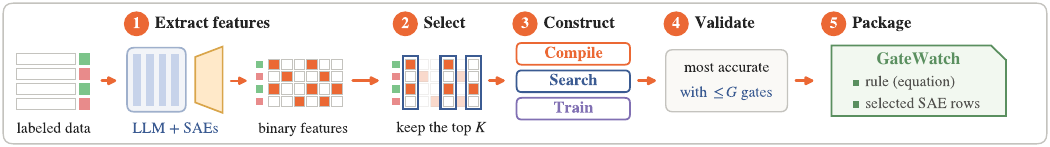}

\caption{\textbf{\bitprobebuild{} Stages.} }
\label{fig:builder}
\label{fig:overview}
\end{figure*}

\para{Activation probes}
For each token at each transformer layer, an LLM produces a hidden vector $\mathbf h\in\mathbb R^d$, called a residual-stream state \citep{elhage2021mathematical}. An \emph{activation probe} is a classifier, typically logistic regression or a small MLP, trained to predict a label from these vectors without modifying the LLM \citep{alain2017probes,belinkov2022probing}. Probes are used to detect misuse and jailbreaks \citep{kramar2026production,cunningham2026constitutional}, while LLM safety testing and attack detection are active areas of systems research \citep{lin2026jailbreak,li2026adr,kuchnik2023relm}. However, individual values in these hidden vectors rarely correspond to recognizable concepts, making it difficult to interpret a probe's predictions from its weights \citep{elhage2022superposition}.

\para{Sparse autoencoders}
A sparse autoencoder (SAE) transforms an LLM's hidden vector into a larger set of learned features, many of which can be associated with human-readable concepts \citep{bricken2023monosemanticity,gao2024scaling}. Its encoder maps the hidden vector $\mathbf h\in\mathbb R^d$ to a feature vector $\mathbf z\in\mathbb R^m$, where $m\gg d$. Although the feature vector is larger, only a small number of features are active for each token, allowing the SAE to reconstruct the original hidden vector from these features. Not all learned features have clear interpretations \citep{templeton2024scaling}. Since an SAE is typically trained for a specific transformer layer, extracting features from multiple layers requires a separate encoder for each layer \citep{lieberum2024gemmascope}. We denote the SAE feature vector for token $t$ at layer $\ell$ by $\mathbf z^{(\ell)}_t$.

\subsection{Classifiers over SAE Features}
\label{sec:bg-additive}

\para{From token features to a prediction}
A common approach trains a classifier using SAE features extracted from a single transformer layer. Since each token has its own feature vector, these vectors are first combined into a single representation, either by selecting the last token or by taking the maximum or sum of each feature across all tokens \citep{gallifant2025saeclassification}. A typical classifier is logistic regression with $L_1$ regularization, which encourages most weights to be zero. Each nonzero weight corresponds to an SAE feature, allowing its contribution to be associated with a human-readable description \citep{kantamneni2025saeprobes,bricken2024features}. However, accuracy depends on the chosen layer, token aggregation method, and SAE size, and SAE-based probes do not consistently outperform probes trained directly on raw activations \citep{kantamneni2025saeprobes,smith2025negative}.

\para{Additive predictions}
Logistic regression computes a prediction score by summing weighted feature values, with each feature contributing independently to the score. This limits which feature interactions it can represent: for example, a linear classifier cannot express XOR over two binary features, where exactly one must be active. Moreover, when many features have nonzero weights, the decision rule becomes difficult to summarize. Nonlinear classifiers can capture more complex interactions, but may also be difficult to interpret.

\para{Serving costs}
SAE encoders can consume substantial GPU memory, especially when features are extracted from multiple layers. Running these encoders on another device saves GPU memory but introduces activation transfers and additional computation. Keeping them on the GPU instead competes with the key--value (KV) cache, which stores attention information from processed tokens \citep{kwon2023vllm}. Under a fixed GPU memory budget, larger encoders leave less space for the KV cache, potentially limiting the number or length of requests served concurrently. We quantify this overhead in \cref{sec:serving}.

\subsection{Boolean Classifiers and Circuit Construction}
\label{sec:bg-search}

\para{Boolean classifiers}
\label{sec:bg-logic}
A Boolean classifier maps binary inputs $b_1,\ldots,b_K$ to a binary prediction using logic operations such as \textsc{and}, \textsc{or}, \textsc{not}, and \textsc{xor}. For example, $(b_1\lor b_2)\land\neg b_3$ predicts positive when at least one of the first two conditions holds and the third does not. Such classifiers can be represented as circuits of connected logic gates. Differentiable logic networks learn these circuits directly through gradient-based training \citep{petersen2022difflogic,petersen2024convolutional}.

\para{Symbolic search}
Symbolic regression searches for expressions that fit training data while favoring simpler expressions \citep{koza1992genetic,espejo2010survey}. When the inputs are binary and the allowed operations are Boolean gates, this approach searches directly for small Boolean classifiers.

\para{Logic synthesis}
In digital hardware design, logic synthesis tools convert functional descriptions of a hardware unit design into optimized networks of logic gates while preserving their behavior. Tools such as Yosys and ABC are widely used for this purpose \citep{wolf2013yosys,brayton2010abc}. We utilize these hardware-design tools to convert trained classifiers with binary inputs and outputs into compact Boolean circuits.

\section{\sys Design}
\label{sec:method}

\sys builds and serves interpretable LLM probes we call \bitprobes. Each \bitprobe is a compact logical rule over interpretable features. \bitprobebuild{} constructs a \bitprobe{} from labeled examples (\cref{SS:design:builder,SS:design:buildertechniques}), and \bitprobeserve runs many deployed \bitprobes together during LLM inference (\cref{SS:design:serv}).

\subsection{\bitprobebuild{}: Building a \bitprobe{}}
\label{SS:design:builder}



\begin{algorithm}[t]
\caption{\bitprobebuild{}}
\label{alg:builder}
\begin{algorithmic}[1]
\REQUIRE LLM and pretrained SAEs; fit and validation data $\mathcal D_{\mathrm{fit}},\mathcal D_{\mathrm{val}}$; range of feature counts $\mathcal K$; gate budget $G$
\ENSURE Best \bitprobe{} with at most $G$ gates

\STATE $B_{\mathrm{fit}},B_{\mathrm{val}}\gets\textsc{BinaryFeatures}(\text{LLM},\text{SAEs},\mathcal D_{\mathrm{fit}},\mathcal D_{\mathrm{val}})$ \label{alg:builder:extract}\label{alg:builder:binary}
\STATE $\mathcal P\gets\varnothing$

\FORALL{$K\in\mathcal K$} \label{alg:builder:configs}
  \STATE $J\gets\textsc{SelectFeatures}(B_{\mathrm{fit}},K)$ \label{alg:builder:features}

  \FORALL{$T\in\{\textsc{Compile},\textsc{Search},\textsc{Train}\}$}
    \STATE $\mathcal Q\gets T(B_{\mathrm{fit}}[:,J],B_{\mathrm{val}}[:,J],G)$ \label{alg:builder:construct}
    \STATE $\mathcal P\gets\mathcal P\cup\{(C,J)\mid C\in\mathcal Q,\ |C|_{\mathrm{gates}}\le G\}$ \label{alg:builder:feasible}
  \ENDFOR
\ENDFOR

\STATE $(C^\star,J^\star)\gets\arg\max_{(C,J)\in\mathcal P}\Acc(C;B_{\mathrm{val}}[:,J])$ \label{alg:builder:select}
\STATE \textbf{return} $\textsc{Package}(C^\star,J^\star)$ \label{alg:builder:used}\label{alg:builder:package}

\end{algorithmic}
\end{algorithm}

\para{Overview}
\Cref{fig:builder} shows the \bitprobebuild{} in five stages. It runs labeled examples through the LLM and its SAEs and turns the SAE activations into binary features \circled{1}, keeps the most informative of them \circled{2}, constructs candidate rules with three techniques \circled{3}, selects the most accurate candidate within the size budget on validation data \circled{4}, 
and packages it as a \bitprobe{} \circled{5}.

\para{Inputs and output}
\Cref{alg:builder} describes \bitprobebuild{} end to end. It takes an LLM with pretrained SAEs, labeled examples split into fit and validation sets, a range $\mathcal K$ of feature counts, and a gate budget $G$, the maximum number of two-input gates in the rule. It turns every example into binary features (\cref{alg:builder:extract}). For each $K\in\mathcal K$, it keeps the $K$ most informative binary features (\cref{alg:builder:features}) and applies all three construction techniques to generate candidate rules over them (\cref{alg:builder:construct}). Among all candidates within the budget (\cref{alg:builder:feasible}), it selects the one with the highest validation accuracy (\cref{alg:builder:select}). The output is a \bitprobe{}: the rule, the meaning of its inputs, and each input's SAE encoder row, bias, and threshold (\cref{alg:builder:package}).

\para{Binary features}
\label{sec:features}
A feature \emph{fires} when its SAE activation is above a threshold. At layer $\ell$ and token $t$ of input $x$, the firing bit of feature $i$ is $b^{(\ell)}_{t,i}(x)=\ind\big[z^{(\ell)}_{t,i}(x)>\tau_i\big]$, where $\tau_i$ is the feature's threshold, provided by the SAE. Computing a firing bit needs only the feature's encoder row, bias, and threshold. \textsc{BinaryFeatures} (\cref{alg:builder:extract}) runs each example through the LLM once, records which features fire at which tokens, and computes three yes/no values per feature: \textsc{Last} (fires at the final token), \textsc{Any} (fires anywhere), and \textsc{Count2} (fires at least twice). We call each value a \emph{binary feature} and write $\LAST{\ell}{i}$, $\ANY{\ell}{i}$, and $\CNT{\ell}{i}$. The latter two capture occurrences that a last-token probe would miss. Later, during serving, the same values are computed as tokens arrive (\cref{SS:design:serv}).

\para{Feature selection}
 A model with $L$ SAE layers of $m$ features yields $3Lm$ binary features, typically millions. Searching over so many inputs is impractical, and a readable rule should use only a few. \textsc{SelectFeatures} (\cref{alg:builder:features}) therefore ranks binary features on the fit set by \emph{firing asymmetry},
\begin{equation}
 r_j=\big|\Pr(b_j=1\mid y=1)-\Pr(b_j=1\mid y=0)\big|,
\label{eq:firing_asymmetry}
\end{equation}
and keeps the top $K$. The fit set decides which features rank highest, and the validation set decides how many to keep.

\para{Constructing and selecting a rule}
For each $K$, every technique returns a set of candidate rules, possibly empty (\cref{alg:builder:construct}). \bitprobebuild{} keeps candidates with at most $G$ gates (\cref{alg:builder:feasible}) and selects the most accurate one on validation data across all techniques and values of $K$ (\cref{alg:builder:select}). \textsc{Package} (\cref{alg:builder:package}) then keeps only the features the rule uses and stores, for each, its layer, SAE index, token rule, encoder row, bias, and threshold. Feature descriptions from Neuronpedia~\citep{lin2024neuronpedia} are attached afterwards and do not influence selection.

\subsection{Construction Techniques}
\label{SS:design:buildertechniques}

\begin{table}[]
\centering
\caption{\textbf{Construction techniques.}}
\label{tab:builders}
\small
\setlength{\tabcolsep}{5pt}
\begin{tabular}{@{}lp{0.72\columnwidth}@{}}
\toprule
Technique & Strength and limitation \\
\midrule
Compile & Reuses a trained classifier through exact logic synthesis;
rounding and pruning may reduce accuracy. \\
\addlinespace[2pt]
Search &  Explores diverse rule structures without a fixed
topology; search becomes costly with many inputs. \\
\addlinespace[2pt]
Train & Scales to more inputs; learns via differentiable gates;
limited by fixed topology and discretization. \\
\bottomrule
\end{tabular}
\end{table}

\Cref{tab:builders} compares the three techniques:
Compile converts a trained classifier into a circuit, Search explores
Boolean expressions, and Train learns logic gates through optimization.
They share the same inputs and size budget; \bitprobebuild{} selects
the most accurate feasible rule on validation set data.

\para{Technique 1: Compile}
Compile turns a trained classifier into a small rule using hardware logic synthesis. This technique was inspired by hardware design, where a computation becomes much smaller when some of its inputs are known in advance. If we synthesize a general multiplier $a\times b$ into logic gates, the circuit must handle every possible value of both $a$ and $b$, so it needs many gates. If $b$ is always the same constant, the synthesis tool builds that constant into the logic, and the circuit for $a\times\mathrm{const}$ needs far fewer gates; multiplying by five, for example, becomes a single shift and an addition. Synthesis tools find such simplifications automatically. A trained classifier offers the same opportunity: its weights are fixed, and only its input features change. Because its inputs are binary features and its output is a single bit, the classifier is a Boolean function, and synthesis can turn it into a gate network with the weights built in. Synthesis itself is exact; the steps around it, rounding the weights and pruning gates, trade a small amount of accuracy for a much smaller rule. \Cref{alg:bitprobe} lists the steps, which we describe in order.

\emph{Fitting and rounding} (\crefrange{alg:line:lr}{alg:line:int}). \Cref{alg:line:lr} fits an $L_1$-regularized logistic classifier over the $K$ selected binary features:
\begin{equation}
 F_{\boldsymbol\beta}(b_1,\ldots,b_K)=\ind\Big[\beta_0+\textstyle\sum_{i=1}^K\beta_i b_i\ge0\Big],
\label{eq:specialized_lr_function}
\end{equation}
where $\beta_0$ is the bias and $\beta_i$ is the learned weight of feature $i$. Because each $b_i$ is a bit, the product $\beta_i b_i$ is either the fixed weight $\beta_i$ or zero, so the rule needs neither a multiplier nor weights supplied as inputs. \Cref{alg:line:int} multiplies the weights and the bias by a shared positive scale $S$ and rounds them to integers $q_i$, which turns the decision into an integer sum and a comparison.

\emph{Synthesis} (\crefrange{alg:line:synth}{alg:line:init}). \Cref{alg:line:synth} passes this integer function to logic synthesis (\cref{sec:bg-search}), with arithmetic widths chosen to avoid overflow. The synthesis tools propagate the constant weights, remove redundant operations, and share repeated computations. The resulting gate network $C$ makes exactly the same predictions as the integer classifier, and \cref{alg:line:init} records it as the first candidate.

\begin{figure}[t]
\centering
\includegraphics[width=\columnwidth]{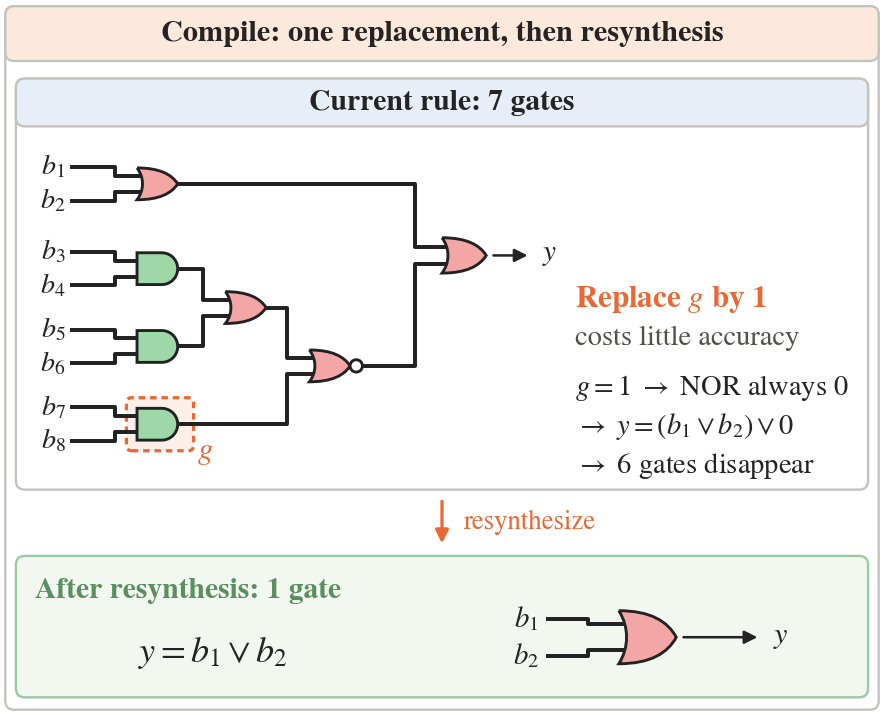}
\vspace{-15pt}
\caption{\textbf{One replacement in Compile.}
Replacing gate $g$ with 1 barely affects validation accuracy.
The NOR gate then always outputs 0, making the final OR
$(b_1\lor b_2)\lor0$. Resynthesis removes the redundant gates,
leaving just $b_1\lor b_2$. The circuit shrinks from 7 gates
to 1 (\cref{alg:line:score,alg:line:apply,alg:line:resynth}).}

\label{fig:pipeline}
\end{figure}

\emph{Pruning to the size budget} (\crefrange{alg:line:loop}{alg:line:add}). Synthesis preserves the function, so it cannot trade accuracy for size; pruning makes this trade explicitly, guided by validation accuracy. While $C$ has more than $G$ gates (\cref{alg:line:loop}), \cref{alg:line:score} examines each gate $g(a,b)$ and
tries replacing its output with zero, one, or either input.
It records the replacement that preserves the most
validation accuracy and how much accuracy is lost. The gate's importance is the smallest validation-accuracy loss among these choices:
\begin{equation}
 I(g)=\Acc(C;\mathcal D_{\mathrm{val}})-\max_{r\in\{0,1,a,b\}}\Acc(C_{g\leftarrow r};\mathcal D_{\mathrm{val}}),
\label{eq:gate_importance}
\end{equation}
where $\Acc$ denotes accuracy on the validation examples $\mathcal D_{\mathrm{val}}$ and $C_{g\leftarrow r}$ replaces the gate's output by $r$; negative importance means that the replacement improves accuracy. \Cref{alg:line:apply} uses \textsc{ReplaceBatch} to apply the least
important replacements in one batch. The parameter $\rho$ limits how many gates can be
replaced per iteration (as a fraction of the circuit), while
$\epsilon$ allows including replacements whose validation accuracy
is close to that of the best single replacement. The result is
then resynthesized (\cref{alg:line:resynth}). \Cref{fig:pipeline} illustrates this effect: replacing a single
gate with the constant 1 makes an entire branch redundant,
allowing resynthesis to reduce the circuit from seven gates
to just one. Thus, pruning one gate can eliminate several
others through resynthesis. If the rule did not shrink, the loop stops (\cref{alg:line:stop}); otherwise \cref{alg:line:add} keeps the smaller rule as a new candidate, and the next pass scores its gates again. Compile returns all recorded rules with at most $G$ gates
(\cref{alg:line:select}).

\begin{algorithm}[t]
\caption{Compile: construction technique}
\label{alg:bitprobe}
\begin{algorithmic}[1]
\REQUIRE Fit and validation sets on binary features; gate budget~$G$
\ENSURE Candidate circuits with at most $G$ gates
\STATE \textbf{Hyperparameters:} scale $S$, pruning fraction $\rho$, tolerance $\epsilon$
\STATE $\boldsymbol\beta\gets\textsc{FitL1Logistic}(\mathcal D_{\mathrm{fit}})$ \label{alg:line:lr}
\STATE $q_i\gets\operatorname{round}(S\beta_i)$ \label{alg:line:int}
\STATE $C\gets\textsc{Synthesize}\big(\ind[q_0+\sum_i q_i b_i\ge0]\big)$ \label{alg:line:synth}
\STATE $\mathcal P\gets\{C\}$ \label{alg:line:init}
\WHILE{$|C|>G$} \label{alg:line:loop}
  \FORALL{gates $g(a,b)$ in $C$}
    \STATE $(I_g,r_g)\gets\textsc{ScoreGate}(C,g,\mathcal D_{\mathrm{val}})$ \label{alg:line:score}
  \ENDFOR
  \STATE $\widehat C\gets\textsc{ReplaceBatch}(C,\{(I_g,r_g)\}_g,\rho,\epsilon)$ \label{alg:line:apply}
  \STATE $C'\gets\textsc{Resynthesize}(\widehat C)$ \label{alg:line:resynth}
  \IF{$|C'|\ge|C|$} \label{alg:line:stop}
    \STATE \textbf{break}
  \ENDIF
  \STATE $C\gets C'$; $\mathcal P\gets\mathcal P\cup\{C\}$ \label{alg:line:add}
\ENDWHILE
\STATE \textbf{return} $\{C\in\mathcal P:|C|\le G\}$ 
\label{alg:line:feasible}\label{alg:line:select}
\end{algorithmic}
\end{algorithm}

\para{Technique 2: Search}
Search uses symbolic regression (\cref{sec:bg-search}) to construct Boolean rules. We restrict its operators to $\{\land,\lor,\oplus,\neg\}$ and its inputs to the $K$ selected binary features, keeping the search tractable and its outputs executable by \bitprobeserve{}. Search produces candidate formulas of different sizes, and \bitprobebuild{} selects the most accurate one within the gate budget using validation data. Thus, the budget controls the tradeoff between rule size and accuracy.

\para{Technique 3: Train}
Given the selected binary features and labels from the fit
set, we construct a differentiable logic gate network
(\cref{sec:bg-logic}). The network has decreasing layer
widths ($H \to H/2 \to H/4 \to \cdots \to 1$), ending
with a single output. Each node has 16 learnable weights,
one for each possible Boolean function of two inputs,
and computes their weighted combination.
We optimize these weights using the labeled examples,
then select the highest-weighted gate at each node
(discretization) to obtain a Boolean circuit.

To improve accuracy, we initialize nodes as pass-through
wires rather than random gates
\citep{petersen2024convolutional}, and use two
regularization terms to encourage discrete gate choices
and reduce the gap between soft and discrete predictions.
\bitprobebuild{} selects the discrete circuit with the
highest validation accuracy, subject to the gate budget.


\subsection{\bitprobeserve: Serving \bitprobes{}}
\label{SS:design:serv}

\begin{figure*}[t]
\centering
\includegraphics[width=\textwidth]{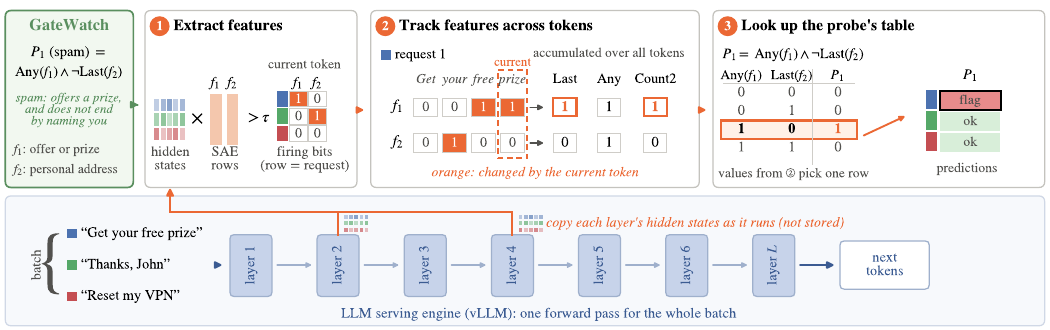}
\vspace{-15pt}

\caption{\textbf{\bitprobeserv{}.} The LLM runs the batch once, and its hidden states stream to the \bitprobe{} as each layer runs. Request~1, ``Get your free prize'': \circled{1} the current token ``prize'' is an offer, so $f_1$ fires, and it is not a personal address, so $f_2$ does not; \circled{2} across tokens, ``free'' and ``prize'' were offers (Any$(f_1){=}1$), and ``your'' was an address but not the last token (Last$(f_2){=}0$); \circled{3} these two bits pick one row of $P_1$'s implementation table, and the request is flagged as spam. Request~2, ``Thanks, John'', ends with a name, so Last$(f_2){=}1$ \circled{1} green row, and it is not flagged.}
\label{fig:serving-design}
\end{figure*}

\Cref{fig:serving-design} depicts \bitprobeserve's execution alongside the LLM. 
\bitprobeserve serves multiple \bitprobes concurrently
in three stages:
extracting SAE features, tracking their occurrences across tokens, and evaluating the probes' lookup tables. Before inference, it collects the features needed by all \bitprobes, groups them, and removes duplicates so each feature is computed only once. It also stores which features each probe uses and its lookup table. This allows \bitprobes to share feature extraction while evaluating different rules. Similar to systems that batch multiple adapters on one LLM \citep{sheng2024slora,chen2024punica}, our approach shares computation across probes, but evaluates Boolean lookup tables instead of adapter weight matrices.

\para{Feature extraction and token rules}
At each selected layer, stage~\circled{1} projects the current hidden states onto the required SAE encoder rows, applies biases and thresholds, and packs the resulting firing bits. Each feature is computed once and shared across probes. Thus, extraction cost depends on the number of distinct features, not the full SAE width or probe count. Stage~\circled{2} updates the token rules for each request: \textsc{Last} tracks firing at the latest token, \textsc{Any} whether the feature has fired so far, and \textsc{Count2} whether it has fired at least twice. We call this \emph{layer streaming}: hidden states are processed as their layer executes, without buffering them for later evaluation. Only firing bits and small counters persist.

\para{Batched lookup evaluation}
Stage~\circled{3} replaces gate-by-gate execution with a lookup table. We convert each Boolean circuit into a truth table: for every possible
binary input, we compute the circuit's output (0 or 1) and store it
at the position indexed by that input. For example, the rule
$(a\land\neg b)\lor c$ has three inputs, so its outputs for
$000,001,\ldots,111$ form the 8-bit table $01011101$ stored in a single byte.
During inference, the observed input bits directly index this table or byte,
returning the circuit's prediction without executing any gates. A single-output circuit with at most $G$ two-input gates
depends on at most $G+1$ distinct binary inputs, so its table requires at most $2^{G+1}$ bits; for $G\le5$, it fits in a single 64-bit word. We batch these lookups across probes and requests, avoiding separate operations for individual gates. Similar lookup-based computation has been used in other systems \citep{wang2025volut}. All three stages run alongside LLM decoding inside the serving engine,
rather than as a separate model, similar to the integration of
structured generation in XGrammar~\citep{dong2025xgrammar}.

\section{Implementation}
\label{sec:impl}

\para{\bitprobebuild{}}
Compile writes the integer probe as a Verilog module, synthesizes it with Yosys \citep{wolf2013yosys} and ABC \citep{brayton2010abc}, and runs the pruning loop of \cref{alg:bitprobe}; gates have two inputs, and inverters on wires are free, as in and-inverter graphs. Search uses PySR \citep{cranmer2023pysr} over $\{\land,\lor,\oplus,\neg\}$ with 600 iterations, 24 populations, and maximum depth 10, fit only on the fit examples. Train uses a DiffLogic network $H \to H/2 \to H/4 \to \cdots \to 1$ trained with Adam (learning rate $10^{-2}$, at most 300 epochs), pass-through initialization, and gate-entropy and soft--hard consistency penalties (0.05 and 0.10).

\para{\bitprobeserve{} runtime}
\bitprobeserve{} is a vLLM plugin \citep{kwon2023vllm} (version 0.30.0) that registers a monitored Gemma-2 model. At every layer that a deployed \bitprobe{} reads, the forward pass multiplies the residual stream by the union of the required encoder rows, thresholds, and bit-packs the result. \textsc{Any} and \textsc{Count2} state is updated in place as bit-planes indexed by vLLM's scheduler slot, so continuous batching and chunked prefill run unchanged. After the last layer, one batched step evaluates all \bitprobes{}: for a probe with truth table $T$ (a 64-bit integer) and input bits $b_1,\ldots,b_m$, $m\le6$, the prediction is $\hat y=(T\gg j)\mathbin{\&}1$ with $j=\sum_i2^{i-1}b_i$. Both stages are plain PyTorch, traced by \texttt{torch.compile} and captured in CUDA graphs with the rest of the model. We also implemented a bit-sliced netlist and a threshold-GEMM evaluator (\cref{app:serving,app:kernels}).

\para{Cost model}
For residual width $d$, a \bitprobe{} with $K_{\mathrm{live}}$ live rows and $G$ gates stores $K_{\mathrm{live}}(2d+8)+4G$ bytes (a BF16 encoder row, bias, and threshold per row, four bytes per gate): 26~KiB for Gemma-2-9B ($d{=}3584$) and the average five-gate \bitprobe{} ($K_{\mathrm{live}}{=}3.8$). Almost all of this is encoder rows: the rule itself takes $4G$ bytes as a gate list, or 8 bytes as the 64-bit table used at serving time, under 0.1\% of the total either way. It costs $K_{\mathrm{live}}d\approx1.4{\times}10^4$ multiply--accumulates per token, versus $md\approx5.9{\times}10^7$ for a 16K-feature encoder at one layer. \textsc{Any} and \textsc{Count2} need one bit or one 2-bit counter of state per feature. Rows are shared across probes, so cost grows with the union of live rows rather than with the number of probes.

\section{Evaluation}
\label{sec:experiments}

We evaluate \sys's ability to create and serve interpretable, accurate \bitprobes, while incurring low overheads.
Specifically, we ask four questions. \textbf{Q1}~How accurate are \bitprobes{} compared with activation and full-SAE probes, and at what memory cost (\cref{sec:results-main})? \textbf{Q2}~How interpretable are \bitprobe's rules (\cref{sec:interpretability-results})? \textbf{Q3}~How efficiently can \bitprobeserve serve multiple \bitprobes running beside an LLM (\cref{sec:serving})? \textbf{Q4}~How do the design choices (layers, token rules, construction technique, gate budget, compilation) trade accuracy against size, and on which tasks does Boolean structure lose accuracy (\cref{sec:ablations})?

\subsection{Methodology}
\label{sec:setup}

\para{Model, datasets and baselines}
We evaluate \sys on Gemma-2-9B \citep{gemma2024gemma2}
using Gemma Scope residual-stream SAEs
\citep{lieberum2024gemmascope}.
We reused the binary classification tasks used 
by \citep{kantamneni2025saeprobes}, henceforth the SAE-Probes dataset, and sampled 
a concept-representative subset of 24 tasks to serve as our evaluation dataset.
We also employ their five
activation-based probes and their full 16K-feature SAE probe as baselines.
We also follow their set split and their use: smaller sets are used for training,
held-out validation data used for hyperparameter selection and candidates,
and accuracy and persistent probe memory (excluding the LLM) are reported on the test set.

\para{Involved layers configuration}
We evaluate three feature-access settings:
\emph{L20} (layer-20 last token), \emph{all-layer} (last token
across 42 layers), and \emph{multi-token} (Last, Any, and Count2
across all layers). Each setting uses Compile, Search, and Train
with gate budgets of 5 and 10. 
The baselines use only layer 20 at the final
token, while our multi-token \bitprobes{} access all
42 layers and tokens. Thus, our main comparison evaluates
deployable systems with different feature access, rather
than classifiers given identical inputs. We separately
evaluate matched feature access in \cref{sec:ablations}.

\para{Setup}
For serving, we run Gemma-2-9B
in BF16 using vLLM 0.30.0 \citep{kwon2023vllm}
on one A100 80GB GPU. All probe baselines use
our vLLM serving plugin,
described in \cref{SS:design:serv}, and we measure throughput and memory under offline and
online workloads. 
We verify \bitprobeserve{} against a PyTorch reference,
checking both kernel implementation as lookup-table and end-to-end
predictions.

Additional dataset and training details are provided in
\cref{app:datasets}, while serving setup and verification
details are provided in \cref{app:serving-setup}.

\begin{figure}[t]
\centering
\includegraphics[width=0.96\columnwidth]{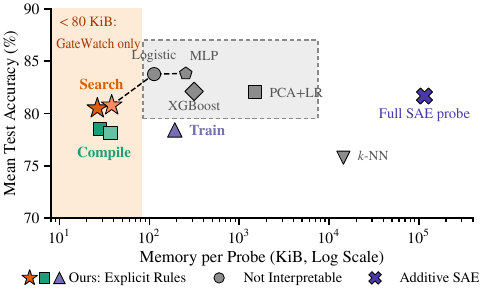}

\caption{\textbf{Accuracy vs.\ memory} (mean over 24 tasks). Search (stars) and Compile (squares) appear twice: dark at $\le$5 gates, light at $\le$10.}
\label{fig:pareto}
\end{figure}

\begin{table}[t]
\centering
\caption{\textbf{Accuracy and memory across 24 tasks.} Features: SAE features the probe uses. W/T/L: tasks won, tied, and lost against the 16K-SAE probe. 
}
\label{tab:aggregate-results}
\small\setlength{\tabcolsep}{3pt}
\begin{tabular}{@{}lrrrcc@{}}
\toprule
Method & Acc.\,\% & KiB & Features & W/T/L & Rule \\
\midrule
Best activation head & 83.84 & 987.4 & -- & 21/0/3 & \ding{55} \\
16K-SAE & 81.66 & 115{,}261 & 16{,}384 & -- & \ding{55} \\
\rowcolor{ours}
Search $\le$5g & \textbf{80.49} & \textbf{26.3} & 3.8 & 10/0/14 & \ding{51} \\
\bottomrule

\end{tabular}
\end{table}

\begin{figure*}[t]
\centering
\includegraphics[width=\textwidth]{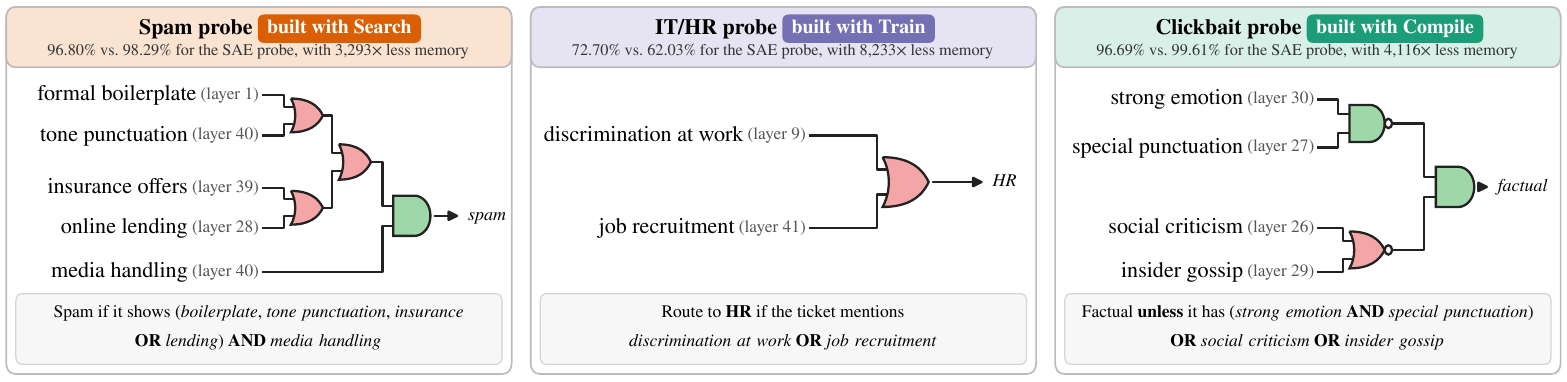}
\vspace{-15pt}
\addtocounter{figure}{1}
\caption{\textbf{Three complete \bitprobes{}}, one per construction technique. Inputs are binary SAE features, named by a paraphrase from Neuronpedia.}
\label{fig:circuits}
\end{figure*}
\addtocounter{figure}{-2}

\subsection{Main Results: Accuracy vs.\ Memory}
\label{sec:results-main}
\label{sec:results-tradeoff}

\para{Accuracy--memory tradeoff}
\Cref{fig:pareto} compares mean test accuracy across
24 tasks against memory per probe (log scale), where
higher accuracy and lower memory are better.
Five-gate Search \bitprobes{} reach 80.49\% accuracy
using 26.3~KiB and 3.8 SAE features on average,
compared with 81.66\% and 115{,}261~KiB for the
16K-SAE probe: a 1.17-point accuracy drop for
4{,}380$\times$ less memory.
This ratio uses the released full 16K-feature encoder; an SAE probe that keeps only the encoder rows of its selected features (265 on average) would need about 1.8~MiB, still 71$\times$ a five-gate Search \bitprobe{}.
Dense activation probes achieve somewhat higher accuracy
but do not provide explicit, readable decision rules.
The best activation probe reaches 83.84\% accuracy,
while the smallest activation baseline, logistic
regression, requires 112.6~KiB, about 4.3$\times$
the memory of a five-gate Search \bitprobe{}.
\Cref{tab:aggregate-results} summarizes the main results.
Five-gate \bitprobes{} also match or outperform the
16K-SAE probe on 10 of 24 tasks.
The gap varies across tasks, and the average gap is not statistically significant: its 95\% confidence interval, $[-3.6, +1.3]$ points, includes zero (Wilcoxon $p{=}0.22$).
Full per-task accuracy and memory results, including the
per-task accuracy differences shown in \cref{fig:pertask}, are provided in
\cref{app:per-task}.

\begin{figure}[!t]
\centering
\includegraphics[width=\columnwidth]{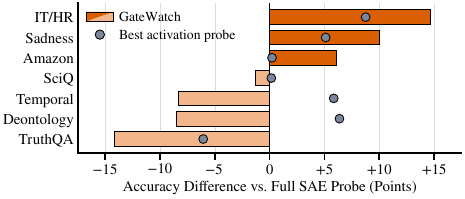}
\vspace{-15pt}
\caption{\textbf{Per-task accuracy difference to the 16K-SAE probe}: three best tasks, one close task, and three worst for five-gate \bitprobes{}.
}
\label{fig:pertask-small}
\end{figure}

\para{The per-task view}
Five-gate \bitprobes{} are the most accurate probes on
8 of 24 tasks, matching or outperforming both
activation-based and SAE-based baselines.
\Cref{fig:pertask-small} illustrates this variation,
showing three strong tasks, one close task, and three
tasks where \bitprobes{} struggle.
For example, \bitprobes{} outperform the best prior
probe on IT/HR ticket routing (76.67\% vs.\ 70.79\%),
Amazon 5-star (95.49\% vs.\ 89.65\%), and tweet
sadness (68.14\% vs.\ 64.07\%).
Full accuracy results across all 24 tasks are shown in
 \cref{fig:pertask} in \cref{app:per-task}.

\addtocounter{figure}{1}

\subsection{Reading Small Probes}
\label{sec:interpretability-results}

Unlike dense classifiers, \bitprobes{} expose their complete decision
rules. 
\Cref{fig:circuits} shows examples from each method.
The Search spam probe combines several possible spam indicators
with a required media-handling feature, achieving 96.80\% accuracy
with 3{,}293$\times$ less memory than the SAE baseline.
The Train IT/HR probe uses a single OR gate over features from
layers 9 and 41, outperforming the SAE baseline by 10.67 points
with 8{,}233$\times$ less memory.
The Compile clickbait probe captures a readable rule involving
emotion, punctuation, criticism, and gossip, reaching 96.69\%
accuracy with 4{,}116$\times$ less memory.
These examples illustrate how small circuits trade some accuracy
for substantial memory savings while making their decisions
explicit. The exact SAE features and their Neuronpedia descriptions
for these three examples are provided in
\cref{tab:circuit-features} in \cref{app:more-circuits}.

\setcounter{topnumber}{1}

\subsection{Serving Efficiency}
\label{sec:serving}

\begin{figure}[t]
\centering
\includegraphics[width=\columnwidth]{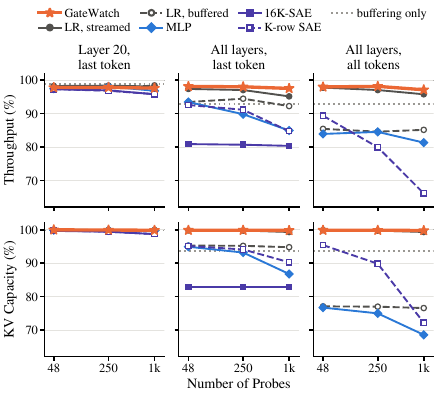}
\vspace{-15pt}
\caption{\textbf{Serving probes inside \bitprobeserve{}.}}
\label{fig:overhead}
\end{figure}

In practice, many probes run alongside the same LLM. We start with 48 real \bitprobes{} and stress-test up to 1{,}000 concurrent probes (\cref{fig:overhead}).
With prior-work access (left), every probe family costs little.
The difference appears as probes read more of the model.
Over all 42 layers (middle), MLP and SAE probes store 1.5--4.6~GiB of weights and keep only 80--85\% of throughput.
Over all layers and tokens (right), pooled MLP and SAE probes also keep 439--588~KiB of state per request and drop to 66--81\% of throughput and 68--72\% of KV capacity.
\bitprobes{} stay at 97--98\% of throughput and at least 99.8\% of KV capacity in every setting, because each reads only a few SAE encoder rows (26--30~MiB for all 1{,}000) and keeps 0.98~KiB of state per request.
LR with our layer streaming comes close (95--96\%) but needs 10$\times$ more weights and gives no readable rule.
The advantage also holds under live request traffic: \bitprobes{} sustain 98.3\% of baseline request throughput, with only a 5.5\% increase in p99 decode latency at 8 requests/s, compared with 23.7\% for MLP and 95.5\% for SAE probes.
Further analyses, including online latency and full measurements, are in \cref{app:serving}.

\subsection{Ablations and Analysis}
\label{sec:ablations}

\begin{figure}[t]
\centering
\includegraphics[width=\columnwidth]{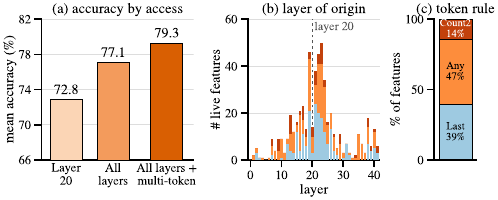}
\vspace{-15pt}
\caption{\textbf{Feature access.} (a)~Mean accuracy over 24 tasks, averaged over the five \bitprobe{} configurations, for each access setting. (b)~Layer of origin of the 527 live features of the 120 multi-token \bitprobes{}, coloured by token rule. (c)~Share of each token rule.}
\label{fig:access}
\end{figure}

\para{A1. Feature access: layers and tokens}
\Cref{fig:access}a shows how accuracy improves as \bitprobes{}
access more of the model. Using only layer 20 at the last token
gives 72.8\% mean accuracy. Accessing all 42 layers raises it
to 77.1\%, and looking across tokens with \textsc{Any} and
\textsc{Count2} raises it further to 79.3\%.
All five \bitprobe{} configurations benefit from both steps.
Where does this improvement come from?
\Cref{fig:access}b shows that the 527 selected features span
40 of 42 layers, with 64\% coming from layers 15--26 and
only 2.7\% from layer 20.
Moreover, \MultiTokenFeatureShare use information across tokens
(47\% \textsc{Any}, 14\% \textsc{Count2}; \cref{fig:access}c).
Detailed accuracy results for all five configurations
and feature-access settings are provided in
\cref{tab:ablation-access} and \cref{fig:ablation},
both in \cref{app:analysis}.

\para{A2. Gate budget: from thousands of gates to five}
We evaluate how restricting the number of gates
affects \bitprobe{} accuracy.
We compare circuits of different 
sizes, from unpruned circuits synthesized from logistic 
regression to \bitprobes{} with five or ten gates.
The unpruned circuits, using only layer 20 at the final
token, have 6{,}157 gates on average and achieve 79.17\%
accuracy. In comparison, five-gate Search \bitprobes{} 
using all layers and tokens achieve 80.49\% accuracy with
39$\times$ less memory. This illustrates that access to 
informative features can compensate for a much smaller circuit.
Reducing Search from ten to five gates lowers accuracy
by only 0.28 percentage points. Additional gate-budget 
comparisons are shown in \cref{fig:ablation} in \cref{app:analysis}.


\para{A3. Construction techniques}
\label{sec:abl-builders}
We compare the accuracy of our three construction techniques
across feature-access settings and gate budgets.
Search is the most accurate on average in every setting (80.49\% at five gates vs.\ 78.48\% for Compile and 78.42\% for Train), but the techniques are complementary: on test data, Search is the most accurate on 12 tasks, Compile on 8 (one tie), and Train on 4. \bitprobebuilder{} builds all three and keeps the validation-best \bitprobe{}.
We also explored combining different
\bitprobes{} through majority voting, learning that their
complementary predictions can improve accuracy.
Detailed results are provided in \cref{tab:logic-ensemble}
in \cref{app:ensemble}.

\para{A4. Boolean structure vs accuracy} 
We evaluate whether more flexible classifiers improve
accuracy when using the same features as \bitprobes{}. We train logistic regression, XGBoost, an MLP, and a
decision tree on exactly the features selected by each five-gate
Search \bitprobe{}. They reach 80.39--80.61\% accuracy, nearly
identical to the \bitprobes{}' 80.49\%.
For Compile and Train,
these classifiers improve accuracy by only 0.5--1.2 points.
Thus, on the features \bitprobe{} selects, the Boolean form costs no accuracy; we have not compared against sparse classifiers that select their own features under the same budget (\cref{sec:limitations}).
Detailed accuracy comparisons with alternative classifiers
are provided in \cref{tab:same-atoms} in \cref{app:analysis}.

\para{A5. Where \bitprobes{} struggle}
We compare performance across content and relational tasks
to understand where \bitprobes{} work well and where they struggle.
On 13 \emph{content} tasks, such as sentiment, spam, and
clickbait detection, where the label depends on which concepts
appear in the text, five-gate \bitprobes{} outperform the SAE
probe by 2.57 percentage points on average.
However, on 11 \emph{relational} tasks, such as NLI,
paraphrase, and question answering, where the model must
understand relationships between sentences or concepts,
they trail by 5.58 points. The largest gaps are on TruthQA
($-14.1$), deontology ($-8.5$), and temporal sense ($-8.3$).

\para{A6. Feature-selection limits}
Selecting too few binary features can make different
examples indistinguishable. For example,
consider two prompts with opposite labels. Using the top
$K=8$ selected features, they produce different binary vectors,
$01010011$ and $01010000$. But if we keep only the top $K=4$
features, both become $0101$. The classifier now sees exactly
the same input for two examples with opposite labels, so it
cannot classify both correctly, no matter how complex it is.
This puts a limit on the best possible accuracy. On tweet sadness,
increasing $K$ from 8 to 128 reduces the fraction of examples
with such collisions from 95\% to 0.2\%, although more test
patterns become unseen during training.
Detailed collision rates and unseen-pattern statistics
across feature counts are provided in
\cref{tab:sadness-collisions} in \cref{app:ensemble}.

\section{Limitations}
\label{sec:limitations}

\para{Interpretability is bounded by the SAE}
A \bitprobe{}'s readability depends on Neuronpedia auto-descriptions,
which are machine-generated, sometimes vague, and occasionally wrong. Thus, even an exact Boolean rule
may be misinterpreted. SAE features can also miss relevant concepts or activate
on irrelevant ones, causing errors even when the circuit is correct.
We have not evaluated whether practitioners can understand and use
these rules more reliably than the top-$k$ coefficients of a linear
probe. Finally, \bitprobe{} requires a pretrained SAE for the
monitored model.
Examples of ambiguous SAE feature descriptions are provided
in \cref{tab:qnli-feature-meanings} in \cref{app:more-circuits}.


\para{Feature selection and access}
We rank binary features individually by firing asymmetry
(\cref{eq:firing_asymmetry}), which may favor \textsc{and}/\textsc{or}
circuits and miss useful feature combinations. Multivariate
feature selection is future work. Prior-work baselines use
layer 20 at the last token, whereas \bitprobes{} access all
layers and tokens. Under matched access, five-gate Search \bitprobes{} trail the SAE probe by 7.64 points (74.02\%). Our controls show
that standard classifiers achieve similar accuracy when given
the same features selected by \bitprobe{}.
However, we have not compared against classifiers that select
their own features from all layers and tokens under the same
memory budget as \bitprobe{}. Detailed matched-feature classifier comparisons are provided
in \cref{tab:same-atoms} in \cref{app:analysis}.

\para{Serving scope}
Since prior-work classifiers are available only for layer 20,
we use randomly initialized baseline classifiers of the same
dimensions at other layers to measure serving overhead.
Experiments with 1000 monitors also include synthetic
\bitprobes{}. These experiments therefore measure serving
cost, not prediction accuracy. All serving experiments use
a single A100 GPU, one model, and one serving system.
Evaluating \bitprobe{} on other hardware, models, and
deployment settings remains future work.
Our accuracy results come from static benchmark classification tasks, not from monitoring live LLM agents for unsafe behavior; evaluating \bitprobes{} on real monitoring traffic is future work.





\section{Related Work}
\label{sec:related}

\para{Probes for LLM monitoring}
Activation probes predict properties of LLM inputs or behavior from internal activations \citep{alain2017probes,belinkov2022probing} and have been used to detect misuse and jailbreaks \citep{kramar2026production,cunningham2026constitutional}. Probes trained on raw activations, such as linear classifiers and small MLPs, can be difficult to interpret because their inputs do not directly correspond to recognizable concepts. In contrast, \bitprobe{} expresses its entire decision rule using interpretable SAE features and explicit Boolean logic. This approach is also relevant to systems for LLM safety testing and attack detection \citep{lin2026jailbreak,li2026adr,kuchnik2023relm}.

\para{Classifiers over SAE features}
SAE features have been used for classification and safety probing \citep{gallifant2025saeclassification,shi2025safer,bricken2024features}. SAE-Probes \citep{kantamneni2025saeprobes} evaluates $L_1$-regularized logistic regression over SAE features across 113 datasets and finds that these probes do not consistently outperform classifiers trained on raw activations; other work reports similar limitations \citep{smith2025negative}. Many existing SAE probes combine features additively, making their decision rules difficult to summarize when many features contribute. PolySAE explores feature interactions by modifying the SAE decoder itself \citep{koromilas2026polysae}. In contrast, \bitprobe{} combines a small number of features from pretrained SAEs across layers and tokens using explicit Boolean logic.

\para{Logic and rule-based classifiers}
Learning Boolean formulas and decision lists is a classical problem \citep{valiant1984theory,rivest1987decision}, while symbolic regression searches for compact expressions that fit data \citep{koza1992genetic,schmidt2009distilling,cranmer2023pysr}. Differentiable logic networks learn combinations of logic gates directly \citep{petersen2022difflogic,petersen2024convolutional}, and binarized networks use low-precision operations for efficient inference \citep{hubara2016binarized,bannink2021larq}. LogicNets and NullaNet explore mapping neural computations to digital hardware circuits \citep{umuroglu2020logicnets,nazemi2019nullanet}. Unlike these approaches, \bitprobe{} constructs small Boolean circuits over named SAE concepts. In addition to direct search and training, our \textsc{Compile} method converts a trained logistic regression probe into a compact circuit using hardware logic synthesis tools, including Yosys and ABC \citep{wolf2013yosys,brayton2010abc}, followed by validation-guided pruning .

\para{Serving many lightweight add-ons}
Existing LLM serving systems efficiently support additional per-request functionality, including LoRA adapters \citep{sheng2024slora,chen2024punica} and grammar-constrained decoding \citep{dong2025xgrammar}, within inference engines such as vLLM \citep{kwon2023vllm,ye2025flashinfer}. Related systems work also explores low-overhead instrumentation of GPU computations \citep{huang2025neutrino}. These techniques motivate efficient integration of lightweight functionality into model serving. \bitprobeserve{} applies this principle to monitoring by sharing SAE feature extraction across probes, streaming binary features across layers without storing full hidden states, and evaluating multiple classifiers in parallel using compact lookup tables.
\section{Conclusion}
\label{sec:conclusion}

\bitprobe{} treats an LLM probe as a Boolean logic object. By binarizing
pretrained SAE features into named atoms, it constructs compact circuits
through three techniques: \textsc{Compile}, which converts trained logistic
probes into Boolean circuits; \textsc{Search}, which searches for compact Boolean expressions;
and \textsc{Train}, which learns circuits through differentiable logic.
The resulting classifiers are simultaneously executable circuits and
human-readable Boolean rules over model concepts. Across 24 tasks, five-gate circuits come within 1.2
points of full-dictionary SAE probes, beat them on 10 tasks, and use
4{,}380$\times$ less memory, 
while \bitprobeserve{}, built on vLLM, enables layer streaming and serves 1{,}000
heterogeneous multi-layer, multi-token circuits at 97--98\% of baseline
throughput.
We see explicit logic over sparse features as a practical bridge between
interpretability research and the memory- and latency-constrained reality of
production LLM monitoring, and identify multivariate atom selection,
pairwise features, and real-time, real-world datasets as next steps.

\ifispublish
\else
    \clearpage
\fi
\draftunboundedpages
\bibliography{Bibs/refs}
\bibliographystyle{mlsys2025}

\section*{AI Usage Disclosure}
\label{sec:ai-disclosure}

We used generative AI tools in the following parts. 
We reviewed, verified, and take responsibility for all content, claims, citations, and line of code.

\para{Writing and editing}
We used Claude to edit drafts: rephrasing sentences, reorganizing sections to fit
the page limit, unifying terminology, and drafting tables. 
All results and conclusions come from the authors' experiments. Claude helped with organizing them. 

\para{Literature search}
We used Claude and ChatGPT to search for related papers. 
We read the abstracts of every suggested paper.

\para{Figures}
We used Claude to write Python scripts that draw the
design figures and the result plots from our result files. The plotted values
come directly from the result files produced by our experiments.

\para{Code}
We used ChatGPT to help write parts
of the experiment and serving code.
All code was reviewed and tested by the authors; correctness was checked as
described in \cref{sec:setup}.

\clearpage
\appendix
\crefalias{section}{appendix}
\crefalias{subsection}{appendix}
\raggedbottom

\renewcommand{\topfraction}{0.9}
\renewcommand{\dbltopfraction}{0.9}
\renewcommand{\textfraction}{0.07}
\renewcommand{\floatpagefraction}{0.8}
\renewcommand{\dblfloatpagefraction}{0.8}
\setcounter{topnumber}{3}
\setcounter{dbltopnumber}{2}
\makeatletter
\setlength{\@fptop}{0pt}
\setlength{\@fpsep}{14pt}\setlength{\@dblfpsep}{14pt}
\setlength{\@fpbot}{0pt plus 1fil}\setlength{\@dblfpbot}{0pt plus 1fil}
\setlength{\@dblfptop}{0pt}
\makeatother

\FloatBarrier
\section{Datasets and Training Details}
\label{app:datasets}

\begin{table}[!htbp]
\centering
\caption{The 24 evaluation tasks. The positive class is the encoded label
         $y{=}1$, which is not always the class suggested by the task name. All
         tasks use 924 fitting and 100 validation examples (TruthQA: 586/100).}
\label{tab:datasets}
\scriptsize\setlength{\tabcolsep}{2.5pt}
\begin{tabular}{@{}lllr@{}}
\toprule
Task & Concept family & Positive class ($y{=}1$) & Test \\
\midrule
CS54 & commonsense & true / yes & 3{,}975 \\
Living room & lexical (sanity) & ``living room'' absent & 2{,}093 \\
SciQ & science QA & correct answer & 3{,}975 \\
QNLI & NLI & non-entailment & 3{,}975 \\
Toxicity & moderation & toxic & 99 \\
Tweet sadness & emotion & sadness & 1{,}475 \\
MNLI & NLI & entailment & 2{,}307 \\
Politician & entity knowledge & is a politician & 3{,}975 \\
CoLA & grammaticality & acceptable & 3{,}975 \\
TruthQA & truthfulness & incorrect answer & 99 \\
Temporal sense & temporal commonsense & plausible & 1{,}199 \\
Physics & physical reasoning & option B correct & 3{,}975 \\
Reasoning & qualitative reasoning & option B correct & 1{,}671 \\
Deontology validity & ethics & excuse valid & 3{,}975 \\
MRPC & paraphrase & equivalent & 1{,}363 \\
QQP & paraphrase & duplicate & 3{,}975 \\
SST-2 & sentiment & positive & 3{,}975 \\
AI-generated text & provenance & AI-generated & 3{,}975 \\
Spam & abuse & spam & 469 \\
Fake news & misinformation & real news & 2{,}095 \\
Clickbait & headline quality & factual (non-clickbait) & 1{,}543 \\
Digestive disease & biomedical & digestive disease & 99 \\
IT/HR support & enterprise routing & HR ticket & 403 \\
Amazon 5-star & review rating & 5-star review & 975 \\
\bottomrule
\end{tabular}
\end{table}

\paragraph{Tasks.}
We chose 24 of the 113 binary SAE-Probes datasets \citep{kantamneni2025saeprobes} so that each probes a different kind of concept: lexical, syntactic, semantic relations (NLI, paraphrase, duplicate questions), factual and scientific knowledge, commonsense, temporal, physical, and qualitative reasoning, ethics, affect, safety and abuse, and domain-specific text. \Cref{tab:datasets} lists each task, its positive class, and its test size.

\paragraph{Splits.}
Following SAE-Probes, each task has a small, balanced training sample of 1{,}024 examples and a larger held-out test set (up to 3{,}975 examples; 99 for three tasks). We split the training sample deterministically into 924 fit and 100 validation examples (TruthQA: 586/100). This mirrors the low-data setting in which probes are usually built.

\paragraph{Baselines.}
We report the five activation classifiers released by SAE-Probes at layer 20 (logistic regression, PCA$+$LR, $k$-NN, XGBoost, and a three-layer MLP), the best of the five per task, and its $L_1$ logistic probe over the full layer-20 16K-feature SAE. All use last-token representations.

\paragraph{Model selection.}
Every hyperparameter is selected on the 100-example validation split, and test labels are used once: the feature count $K$, the Compile candidate, the Train checkpoint, and the Search formula. For the baselines, we select the best activation classifier and the SAE probe's $K$ by the released validation scores. \Cref{app:oracle} compares this with selecting on the test set.

\paragraph{Metrics and memory.}
We report hard-decision test accuracy. Memory excludes the LLM and counts persistent probe state (\cref{sec:impl}). For the 16K-SAE probe we report the full-encoder deployment, 115{,}261~KiB.

\FloatBarrier

\section{Circuit Inputs and Additional Circuits}
\label{app:more-circuits}

\begin{table}[!htbp]
\centering
\caption{Inputs of the circuits in \cref{fig:circuits}. Descriptions are
         Neuronpedia's (not ours).}
\label{tab:circuit-features}
\footnotesize\setlength{\tabcolsep}{2.5pt}
\begin{tabular}{@{}l l p{0.53\linewidth}@{}}
\toprule
& Atom & Neuronpedia description \\
\midrule
\grp{3}{Spam, Search $\le$5g, last-token atoms (96.80\%)} \\
$f_1$ &
\href{https://www.neuronpedia.org/gemma-2-9b/1-gemmascope-res-16k/13309}
{\textsc{Last}
L1:13309} & specific elements related to legal and formal documents, such as
citations, references, and structured content \\
$f_2$ &
\href{https://www.neuronpedia.org/gemma-2-9b/40-gemmascope-res-16k/6830}
{\textsc{Last}
L40:6830} & punctuation and capitalization cues that indicate the tone and
structure of sentences \\
$f_3$ &
\href{https://www.neuronpedia.org/gemma-2-9b/39-gemmascope-res-16k/10512}
{\textsc{Last}
L39:10512} & promotional offers related to insurance or service policies \\
$f_4$ &
\href{https://www.neuronpedia.org/gemma-2-9b/28-gemmascope-res-16k/14295}
{\textsc{Last}
L28:14295} & information related to online lending and borrowing options \\
$f_5$ &
\href{https://www.neuronpedia.org/gemma-2-9b/40-gemmascope-res-16k/10265}
{\textsc{Last}
L40:10265} & code structures and syntax elements related to media types and
media handling in programming \\
\dashrule
\grp{3}{IT/HR support, Train, last-token atoms (72.70\%)} \\
$f_1$ &
\href{https://www.neuronpedia.org/gemma-2-9b/9-gemmascope-res-16k/15919}
{\textsc{Last}
L9:15919} & instances of employment discrimination and related legal terms \\
$f_2$ &
\href{https://www.neuronpedia.org/gemma-2-9b/41-gemmascope-res-16k/11269}
{\textsc{Last}
L41:11269} & phrases related to job recruitment and alerts \\
\dashrule
\grp{3}{CoLA acceptable, Search $\le$5g (80.00\%)} \\
$f_1$ &
\href{https://www.neuronpedia.org/gemma-2-9b/27-gemmascope-res-16k/15112}
{\textsc{Any}
L27:15112} & identification and investigation processes in medical or biological
contexts \\
$f_2$ &
\href{https://www.neuronpedia.org/gemma-2-9b/16-gemmascope-res-16k/6766}
{\textsc{Any}
L16:6766} & data or results related to growth, financial performance, and
scientific findings \\
$f_3$ &
\href{https://www.neuronpedia.org/gemma-2-9b/7-gemmascope-res-16k/85}
{\textsc{Any}
L7:85} & legal terms and references to evidence and legal concepts \\
\dashrule
\grp{3}{Clickbait, Compile $\le$5g (96.69\%; $y{=}1$: factual)} \\
$f_1$ &
\href{https://www.neuronpedia.org/gemma-2-9b/30-gemmascope-res-16k/728}
{\textsc{Any}
L30:728} & strong emotional expressions or impactful statements \\
$f_2$ &
\href{https://www.neuronpedia.org/gemma-2-9b/27-gemmascope-res-16k/14662}
{\textsc{Any}
L27:14662} & punctuation marks and special characters \\
$f_3$ &
\href{https://www.neuronpedia.org/gemma-2-9b/26-gemmascope-res-16k/7275}
{\textsc{Last}
L26:7275} & evaluation or criticism of social concepts \\
$f_4$ &
\href{https://www.neuronpedia.org/gemma-2-9b/29-gemmascope-res-16k/1786}
{\textsc{Any}
L29:1786} & insider information or gossip \\
\bottomrule
\end{tabular}
\end{table}

\begin{figure}[!htbp]
\centering
\includegraphics[width=\linewidth]
{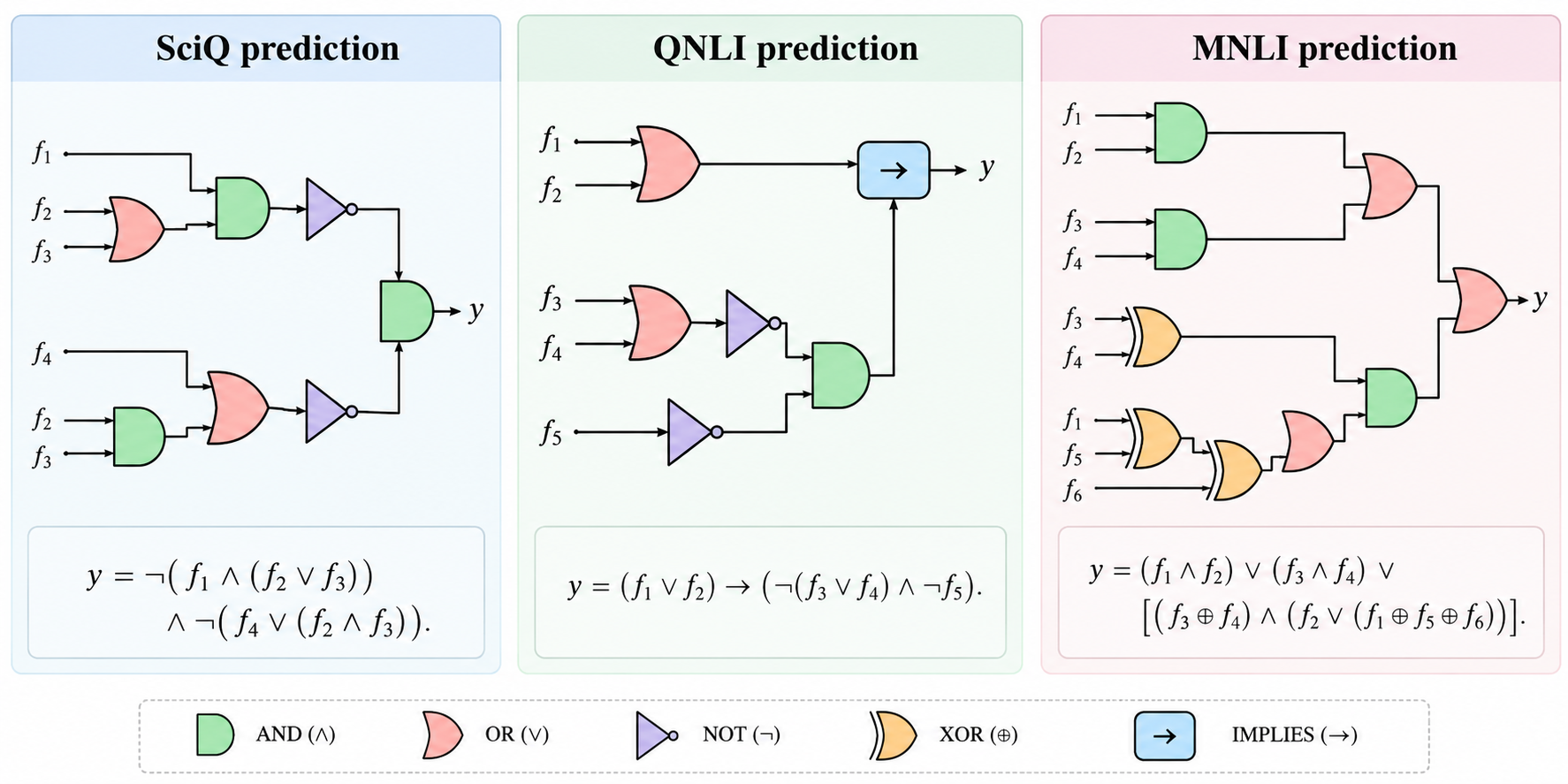}
\caption{Hard Boolean probes for SciQ, QNLI, and MNLI. Inputs are binarized SAE
         features (layer, index); internal nodes are gates; the final node is
         the classifier output.}
\label{fig:interpretable-circuits}
\end{figure}

\Cref{fig:interpretable-circuits} shows three further multi-token circuits from
earlier runs.

\begin{table}[!htbp]
\centering
\caption{Neuronpedia descriptions of the five SAE features in the QNLI circuit.}
\label{tab:qnli-feature-meanings}
\small
\setlength{\tabcolsep}{4pt}
\begin{tabular}{@{}lp{0.15\linewidth}p{0.52\linewidth}@{}}
\toprule
Feature & Atom & Neuronpedia explanation \\
\midrule
\href{https://www.neuronpedia.org/gemma-2-9b/23-gemmascope-res-16k/11132}
{L23:F11132}
& \textsc{Any} & Technological or scientific processes and entities,
particularly calculations and functions. \\
\href{https://www.neuronpedia.org/gemma-2-9b/24-gemmascope-res-16k/5380}
{L24:F5380}
& \textsc{Count2} & Intellectual property and its management in different
contexts. \\
\href{https://www.neuronpedia.org/gemma-2-9b/15-gemmascope-res-16k/488}
{L15:F488}
& \textsc{Last} & Phrases indicating social awareness or activism. \\
\href{https://www.neuronpedia.org/gemma-2-9b/16-gemmascope-res-16k/12222}
{L16:F12222}
& \textsc{Any} & Crisis and emergency terminology, particularly in public-policy
and social contexts. \\
\href{https://www.neuronpedia.org/gemma-2-9b/23-gemmascope-res-16k/7837}
{L23:F7837}
& \textsc{Last} & Punctuation and formatting markers. \\
\bottomrule
\end{tabular}
\end{table}

\FloatBarrier

\section{Validation vs.\ Test-Oracle Model Selection}
\label{app:oracle}

An earlier version of our analysis chose, for every method, the $K$ with the
best \emph{test} accuracy (and the best activation head by test AUC).
\Cref{tab:oracle} compares those optimistic means with the validation-selected
numbers used throughout the paper. Selection on validation costs 0--1.4 points
for Compile and Search, 0.94 points for the SAE probe, and 0.9 points for Train;
the ordering of methods and the gap between five-gate circuits and the SAE probe
(0.90 vs.\ 1.17 points) are unchanged.

\begin{table}[!htbp]
\centering
\caption{Mean test accuracy (\%) over 24 tasks under test-oracle $K$ selection
         and under validation selection (reported everywhere else). W/T/L
         against the 16K-SAE probe selected the same way; MT: multi-token
         atoms.}
\label{tab:oracle}
\scriptsize\setlength{\tabcolsep}{8pt}
\begin{tabular}{@{}lrrrcc@{}}
\toprule
Method & Oracle & Valid. & $\Delta$ & \multicolumn{2}{c}{W/T/L vs.\ SAE} \\
& & & & oracle & valid. \\
\midrule
Best activation head & 84.06 & 83.84 & $-$0.22 & 17/1/6 & 21/0/3 \\
16K-SAE $L_1$ probe & 82.60 & 81.66 & $-$0.94 & -- & -- \\
Exact circuit (L20) & 80.28 & 79.17 & $-$1.11 & 2/1/21 & 4/0/20 \\
Train (DiffLogic), MT & 79.32 & 78.42 & $-$0.90 & 7/0/17 & 8/0/16 \\
Compile $\leq$10g, MT & 79.57 & 78.17 & $-$1.40 & 8/1/15 & 8/0/16 \\
Search $\leq$10g, MT & 82.06 & 80.77 & $-$1.29 & 11/0/13 & 10/0/14 \\
Compile $\leq$5g, MT & 78.48 & 78.48 & \phantom{$+$}0.00 & 7/0/17 & 8/0/16 \\
Search $\leq$5g, MT & 81.70 & 80.49 & $-$1.21 & 11/0/13 & 10/0/14 \\
\bottomrule
\end{tabular}
\end{table}

\FloatBarrier

\section{Earlier Ten-Task Analyses}
\label{app:ensemble}

The analyses in this appendix were run in an earlier configuration (all-layer
last-token atoms, ten tasks, test-oracle $K$); their per-task values therefore
differ from \cref{tab:per-task-full}.

\paragraph{Ensembles.}
The majority vote predicts $\hat y_{\mathrm{maj}}=\ind[\hat y_{\mathrm{LR}}+\hat
y_{\mathrm{PySR}}+\hat y_{\mathrm{DL}}\ge2]$; the any-correct oracle counts an
example as correct if any circuit predicts its label, $\ind[\bigvee_m\hat
y_m(x)=y]$, and is a diagnostic only. \Cref{tab:logic-ensemble} reports both.

The three construction techniques can produce different circuits for the same task, leading to different predictions and errors. We explore whether combining these circuits can improve accuracy in an earlier ten-task experiment. First, we use **majority voting**: each of the three circuits predicts independently, and the final prediction is the one chosen by at least two circuits. This improves accuracy over the best individual circuit on six tasks. We also measure an **any-correct oracle**, which counts an example as correct whenever at least one of the three circuits predicts correctly. Unlike majority voting, this assumes we know which prediction is correct, so it cannot be used in practice. The oracle reaches 91.76\% mean accuracy across ten tasks, indicating that the circuits make complementary errors and that there is potential to improve how their predictions are combined.

\begin{table*}[!htbp]
\centering
\caption{Accuracy (\%) of three independently constructed $\le10$-gate
         circuits, their any-correct oracle, and their deployable majority vote
         (earlier ten-task configuration).}
\label{tab:logic-ensemble}
\small
\setlength{\tabcolsep}{3.5pt}
\begin{tabular}{@{}lrrrrrr@{}}
\toprule
Task & 16K SAE & Compile$\le$10 & Search$\le$10 & Train & Oracle & Majority \\
\midrule
CS54 & 66.14 & 57.76 & 63.95 & 55.27 & 91.92 & 62.24 \\
Living room & 99.38 & 98.09 & 99.47 & 99.86 & 99.95 & 99.52 \\
SciQ & 91.14 & 87.52 & 90.84 & 89.89 & 95.52 & \textbf{90.99} \\
QNLI & 84.30 & 76.98 & 80.60 & 77.84 & 92.88 & \textbf{82.04} \\
Toxicity & 87.88 & 81.82 & 86.87 & 83.84 & 94.95 & \textbf{89.90} \\
Tweet sadness & 63.25 & 59.19 & 65.08 & 65.36 & 85.69 & \textbf{67.53} \\
MNLI & 81.19 & 74.82 & 75.86 & 69.48 & 87.82 & \textbf{76.42} \\
Politician & 69.91 & 62.47 & 67.57 & 62.47 & 85.18 & 66.72 \\
CoLA & 77.46 & 69.31 & 76.15 & 75.22 & 89.74 & \textbf{77.41} \\
TruthQA & 85.86 & 65.66 & 76.77 & 67.68 & 93.94 & 76.77 \\
\midrule
Mean, ten tasks & 80.65 & 73.36 & 78.32 & 74.69 & 91.76 & 78.95 \\
Mean, nine tasks & 79.85 & 72.42 & 77.37 & 73.67 & 91.40 & 77.74 \\
\bottomrule
\end{tabular}
\end{table*}

\paragraph{Label collisions.}
Selecting $K$ atoms maps every example to one of $2^K$ bit patterns, so two
examples with different labels can share a pattern; no deterministic classifier
over those bits, whatever its size or training method, can classify both
correctly. With $n_0(p)$ and $n_1(p)$ the negatives and positives sharing
pattern $p$, the accuracy of the best lookup table is
$\Acc_{\mathrm{pattern}}=\frac1N\sum_p\max\{n_0(p),n_1(p)\}$, a
representation-level ceiling. \Cref{tab:sadness-collisions} shows that raising
$K$ removes collisions but makes most test patterns unseen.

\begin{table}[!htbp]
\centering
\caption{Collision analysis for tweet sadness with last-token atoms. A
         collision pattern occurs with both labels; ``in collisions'' counts
         examples assigned to such patterns.}
\label{tab:sadness-collisions}
\small
\setlength{\tabcolsep}{2.5pt}
\begin{tabular}{@{}rrrrrr@{}}
\toprule
$K$ & Patterns & Collision patt. & In collisions & Fraction & Test seen \\
\midrule
8 & 200 & 136 & 2,376 & 95.08\% & -- \\
16 & 1,178 & 172 & 1,292 & 51.70\% & 53.63\% \\
32 & 2,119 & 37 & 192 & 7.68\% & 13.29\% \\
64 & 2,263 & 9 & 41 & 1.64\% & 8.07\% \\
128 & 2,289 & 2 & 5 & 0.20\% & 6.78\% \\
\bottomrule
\end{tabular}
\end{table}

\FloatBarrier

\section{Circuit-Execution Implementations}
\label{app:kernels}

Before building the vLLM runtime of \cref{SS:design:serv}, we evaluated seven
standalone implementations of hard Boolean circuits: (1)~\emph{naive PyTorch},
one tensor op per gate, launch-bound; (2)~\emph{packed Python}, many examples
per arbitrary-precision integer, still interpreter-bound; (3)~\emph{generated
Python}, straight-line functions without a graph interpreter; (4)~\emph{compiled
C} with 64 examples per word, which removes Python overhead and lets the
compiler inline and reuse registers; (5)~\emph{sample-wise Triton}
\citep{tillet2019triton}, one kernel that interprets gate types from an array;
(6)~\emph{packed-32 Triton}, 32 examples per word, where packing and irregular
feature gathers dominate for very small circuits; and (7)~\emph{generated CUDA},
the fused, ahead-of-time compiled kernel used in the standalone measurements
(\cref{tab:standalone-kernel}). Implementations (1)--(6) remained dominated by
per-gate launches, interpretation, or packing overhead. The generated-CUDA
backend emits straight-line \texttt{AND}/\texttt{OR}/\texttt{XOR}/\texttt{NOT}
code for the complete collection of monitors, compiled ahead of time to PTX and
a device-specific \texttt{.cubin}, so the compiler can propagate constants,
reuse registers, and simplify shared subexpressions across circuits; each thread
handles one example, loads the union of required firing bits, evaluates every
circuit in registers, and writes the packed predictions once. Inside vLLM we
instead use the PyTorch evaluators of \cref{app:serving}, which are captured in
CUDA graphs with the model and need no separate compilation step.

\begin{table}[!htbp]
\centering
\caption{Standalone A100 classifier-head benchmark ($\mu$s). Speedup is MLP
         time over fused-logic time.}
\label{tab:standalone-kernel}
\small\setlength{\tabcolsep}{3pt}
\begin{tabular}{@{}rrrrrrr@{}}
\toprule
& \multicolumn{3}{c}{Batch 1} & \multicolumn{3}{c}{Batch 1,024} \\
\cmidrule(lr){2-4}\cmidrule(lr){5-7}
Mon. & MLP & Logic & Speedup & MLP & Logic & Speedup \\
\midrule
50 & 164.90 & 32.69 & 5.04$\times$ & 495.40 & 34.74 & 14.26$\times$ \\
100 & 153.61 & 23.49 & 6.54$\times$ & 865.58 & 51.48 & 16.81$\times$ \\
200 & 153.57 & 23.43 & 6.56$\times$ & 1622.73 & 76.51 & 21.21$\times$ \\
\bottomrule
\end{tabular}
\end{table}

\Cref{tab:standalone-kernel} compares the standalone fused kernel with matched
MLP heads on an A100, isolating monitor-head execution at one captured layer:
the fused kernel is 5--7$\times$ faster at batch one and 14--21$\times$ faster
at batch 1{,}024.

\FloatBarrier

\section{Additional Ablations}
\label{app:analysis}

\begin{figure*}[!htbp]
\centering
\includegraphics[width=\textwidth]
{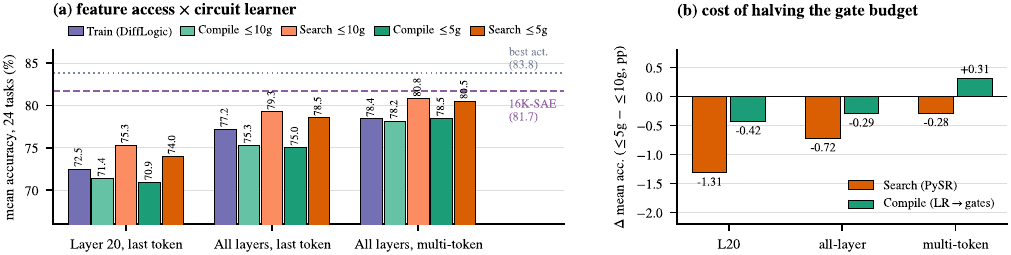}
\caption{\textbf{Ablations} (validation-selected). (a)~Mean accuracy for every
         feature-access setting and technique; dashed and dotted lines are the
         16K-SAE probe and the best activation head. For every technique, each
         step (layer 20 $\to$ all layers $\to$ multi-token) helps. (b)~Cost of
         halving the size budget from ten to five gates: at most 0.3 points with
         multi-token binary features.}
\label{fig:ablation}
\end{figure*}

\Cref{tab:ablation-access} lists the mean accuracy of every \bitprobe{}
configuration in the three feature-access settings, and \cref{fig:ablation}
plots the same results together with the cost of halving the size budget.

\begin{table}[!htbp]
\centering
\caption{\textbf{Full feature-access ablation} (validation-selected). Mean
         accuracy (\%) over 24 tasks and number of tasks where the \bitprobe{}
         is $\ge$ the 16K-SAE probe (``$\ge$S''). L20 reads layer 20 at the last
         token; all-layer reads all 42 layers at the last token; multi-token
         adds \textsc{Any} and \textsc{Count2}. Last column: multi-token minus L20.}
\label{tab:ablation-access}
\footnotesize\setlength{\tabcolsep}{2.0pt}
\begin{tabular}{@{}l rc;{2pt/2pt}rc;{2pt/2pt}rc r@{}}
\toprule
& \multicolumn{2}{c;{2pt/2pt}}{L20} & \multicolumn{2}{c;{2pt/2pt}}{All-layer} &
\multicolumn{2}{c}{Multi-token} & \\
Technique & Acc. & $\ge$S & Acc. & $\ge$S & Acc. & $\ge$S & $\Delta$ \\
\midrule
Train & 72.51 & 1 & 77.20 & 6 & 78.42 & 8 & $+$5.91 \\
Compile $\leq$10g & 71.37 & 0 & 75.30 & 3 & 78.17 & 8 & $+$6.80 \\
Search $\leq$10g & 75.33 & 2 & 79.26 & 8 & 80.77 & 10 & $+$5.45 \\
Compile $\leq$5g & 70.95 & 0 & 75.01 & 3 & 78.48 & 8 & $+$7.54 \\
Search $\leq$5g & 74.02 & 1 & 78.54 & 7 & 80.49 & 10 & $+$6.47 \\
\bottomrule
\end{tabular}
\end{table}

\begin{table}[!htbp]
\centering
\caption{\textbf{Same-feature controls} (validation-selected; mean test accuracy
         over 24 tasks, \%). Each classifier receives exactly the live binary features of
         the named five-gate \bitprobe{} (columns); parentheses give the difference
         to the \bitprobe{} itself. Tree: decision tree.}
\label{tab:same-atoms}
\footnotesize\setlength{\tabcolsep}{3pt}
\begin{tabular}{@{}lrrr@{}}
\toprule
Classifier & Search & Compile & Train \\
\midrule
\bitprobe{} ($\le$5g) & \textbf{80.49} & \textbf{78.48} & \textbf{78.42} \\
\midrule
LR & 80.39 ($-$0.10) & 79.27 ($+$0.78) & 79.07 ($+$0.65) \\
XGBoost & 80.51 ($+$0.02) & 79.63 ($+$1.15) & 79.13 ($+$0.71) \\
MLP & 80.61 ($+$0.12) & 79.56 ($+$1.08) & 79.08 ($+$0.65) \\
Tree & 80.43 ($-$0.06) & 79.25 ($+$0.77) & 78.90 ($+$0.48) \\
\bottomrule
\end{tabular}
\end{table}
 
\section{vLLM Serving Details}
\label{app:serving}

\paragraph{Serving setup.}
\label{app:serving-setup}
Every baseline probe runs in the same vLLM plugin as \bitprobe{} (\cref{sec:impl}), with the same taps and CUDA graphs, at the three access levels of \cref{sec:setup}. LR and MLP heads read the residual stream; SAE probes use the full 16K-feature encoder (16K-SAE) or only the encoder rows their probe needs (K-row SAE). For multi-token access, heads read mean-pooled (LR, MLP) or max-pooled (K-row SAE) features over the prompt; a pooled 16K-SAE over 42 layers does not fit in memory (about 10~GiB of activations per step). Only the layer-20 baselines are trained probes; the all-layer baselines use random weights of the correct shape, because their cost does not depend on the weight values. Up to $N{=}48$ \bitprobes{} we serve the real ones; for larger $N$ we add synthetic \bitprobes{} that keep each real one's gates, token rules, and layers and redraw only its SAE features, which conservatively limits row sharing. The offline workload times vLLM's offline generation API (two repeats) with prefix caching off and up to 8{,}192 batched tokens per step; the online workload sends Poisson arrivals at 2--32 requests/s to a vLLM server.

\begin{table}[!htbp]
\centering
\caption{\textbf{Serving 1{,}000 probes in vLLM.} Weights and per-request
         state carried across tokens (``--'': last token only); KV-cache
         capacity and offline output throughput relative to baseline vLLM;
         online saturation throughput (Sat.) and increase in p99 time per
         output token at 8 requests/s ($\Delta$TPOT), both relative to baseline
         vLLM. ``--'' in the online columns: not run.}
\label{tab:serving}
\footnotesize\setlength{\tabcolsep}{1.3pt}
\begin{tabular}{@{}lrrrrrr@{}}
\toprule
& Weights & State & KV & Offl. & Sat. & $\Delta$TPOT \\
System & MiB & KiB/req & \% & \% & \% & \% \\
\midrule
\multicolumn{7}{@{}l}{\emph{Layer 20, last token (prior-work access)}} \\
LR, buffered & 6.8 & -- & 99.9 & 98.5 & 99.0 & $+$2.1 \\
LR, streamed & 6.8 & -- & 99.9 & 98.1 & 99.9 & $+$0.6 \\
MLP & 110 & -- & 99.7 & 96.8 & 99.0 & $+$2.3 \\
16K-SAE & 115 & -- & 98.7 & 95.8 & 97.8 & $+$4.4 \\
K-row SAE & 115 & -- & 98.7 & 95.8 & -- & -- \\
\bitprobe{} & \textbf{26} & -- & \textbf{99.9} & \textbf{97.6} & \textbf{99.4} & \textbf{$+$1.0} \\
\midrule
\multicolumn{7}{@{}l}{\emph{All 42 layers, last token}} \\
LR, buffered & 287 & -- & 94.8 & 92.3 & 95.8 & $+$4.5 \\
LR, streamed & 287 & -- & 99.4 & 95.2 & 97.1 & $+$10.3 \\
MLP & 4{,}594 & -- & 86.8 & 84.9 & 90.3 & $+$16.8 \\
16K-SAE & 4{,}709 & -- & 82.8 & 80.3 & -- & -- \\
K-row SAE & 1{,}541 & -- & 90.3 & 84.8 & 91.1 & $+$11.0 \\
\bitprobe{} & \textbf{30} & -- & \textbf{99.8} & \textbf{97.5} & \textbf{98.7} & \textbf{$+$4.4} \\
\midrule
\multicolumn{7}{@{}l}{\emph{All 42 layers, multi-token}} \\
LR mean, buffered & 287 & 588 & 76.5 & 85.1 & -- & -- \\
LR mean, streamed & 287 & 3.9 & 99.3 & 95.8 & 97.4 & $+$8.4 \\
MLP, mean-pool & 4{,}594 & 588 & 68.5 & 81.3 & 81.0 & $+$23.7 \\
K-row SAE, max-pool & 1{,}541 & 439 & 72.2 & 66.1 & 73.6 & $+$95.5 \\
\bitprobe{} & \textbf{27} & \textbf{0.98} & \textbf{99.8} & \textbf{97.1} & \textbf{98.3} & \textbf{$+$5.5} \\
\bottomrule
\end{tabular}
\end{table}

\paragraph{Layer streaming versus buffering.}
\label{sec:abl-serving}
To isolate layer streaming from the classifier, we run the same LR heads in two designs: \emph{buffered}, which stores the residual stream of every tapped layer and applies all heads at the end, as a hook-based probe would, and \emph{streamed}, which accumulates each head's partial logit (or running mean) layer by layer (\cref{fig:overhead}, dashed vs.\ solid grey). Buffering the 42 residual streams alone already costs 7.2\% of throughput and 6.2\% of KV capacity. At $N{=}1{,}000$, buffered LR over all layers loses 7.7\% of throughput and 5.2\% of KV capacity, and buffered mean-pooled LR loses 14.9\% and 23.5\%. Streaming the same heads recovers most of this loss (95.2\% and 95.8\% of throughput; 99.4\% and 99.3\% of KV capacity) and cuts pooled state from 588 to 3.9~KiB per request.

\begin{figure*}[!htbp]
\centering
\includegraphics[width=\textwidth]
{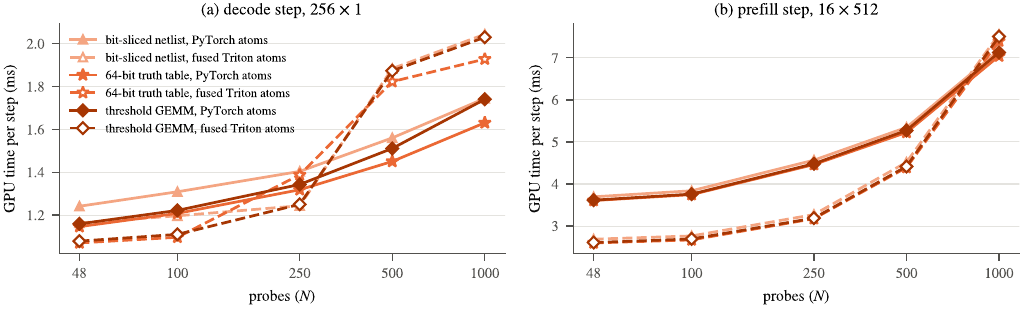}
\caption{\textbf{Circuit evaluators} for multi-token \bitprobe{} banks: probe
         GPU time per step for the three evaluators, with feature extraction in
         PyTorch (solid) or a fused Triton kernel (dashed).}
\label{fig:impls}
\end{figure*}

\paragraph{Circuit evaluators.}
All three evaluators consume the same bit-packed binary features. The \emph{bit-sliced
netlist} merges all circuits into one levelized netlist of two-input gates with
free inverters and packs 32 requests per integer, so each gate is one bitwise
operation for 32 requests. The \emph{truth table} exploits that a five-gate
circuit has at most six live inputs: each circuit is stored as a 64-bit table
indexed by its input bits. The \emph{threshold GEMM} evaluates the circuits as
small integer matrix products followed by thresholds. \Cref{fig:impls} compares
them, with feature extraction in PyTorch or in a fused Triton kernel.

\begin{figure*}[!htbp]
\centering
\includegraphics[width=\textwidth]
{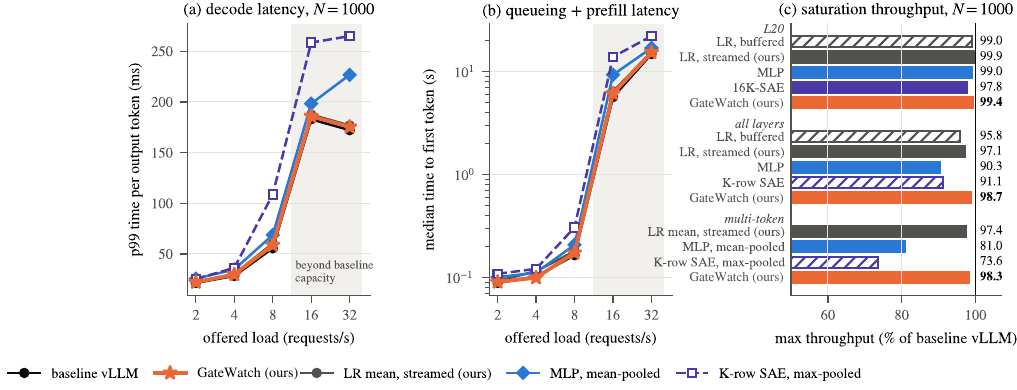}
\caption{\textbf{Online serving} with 1{,}000 probes (Poisson arrivals, 512
         prompt / 128 output tokens). (a,b)~Tail decode latency and median time
         to first token for the multi-token systems; the shaded region lies
         beyond baseline vLLM's capacity (11.2 requests/s). (c)~Saturation
         request throughput for every access level, relative to baseline vLLM;
         hatched bars: buffered LR and K-row SAE.}
\label{fig:online}
\end{figure*}

\begin{figure*}[!htbp]
\centering
\includegraphics[width=\textwidth]
{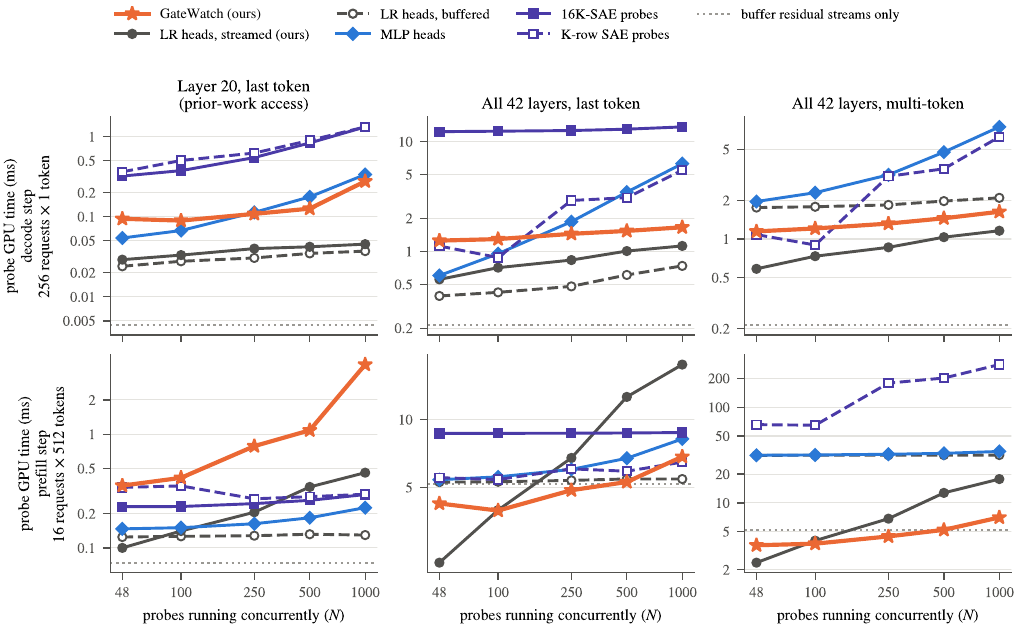}
\caption{\textbf{Probe GPU time per engine step} (CUDA-graph replay,
         log--log), for a decode step of 256 requests (top) and a prefill step
         of 16 requests $\times$ 512 tokens (bottom). Dotted: cost of buffering
         the residual streams alone.}
\label{fig:step-cost}
\end{figure*}

\begin{figure*}[!htbp]
\centering
\includegraphics[width=\textwidth]
{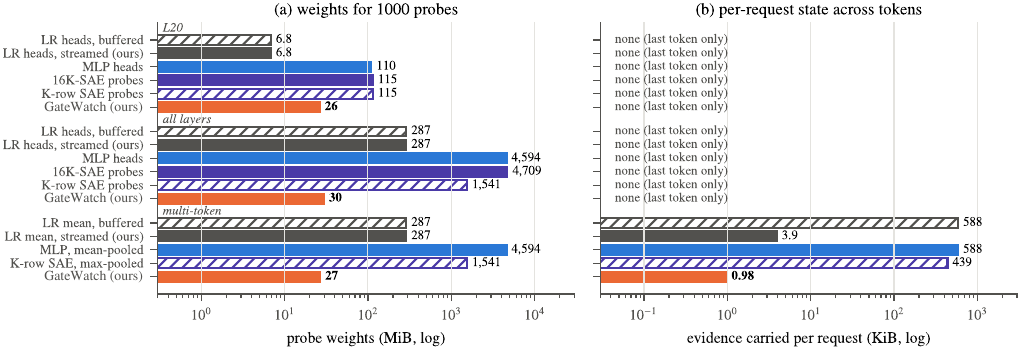}
\caption{\textbf{Probe memory for 1{,}000 probes.} (a)~Weights.
         (b)~Binary features carried per request between tokens, excluding the
         one-byte-per-probe output buffer; last-token systems carry none.}
\label{fig:serving-memory}
\end{figure*}

\begin{figure*}[!htbp]
\centering
\includegraphics[width=\textwidth]
{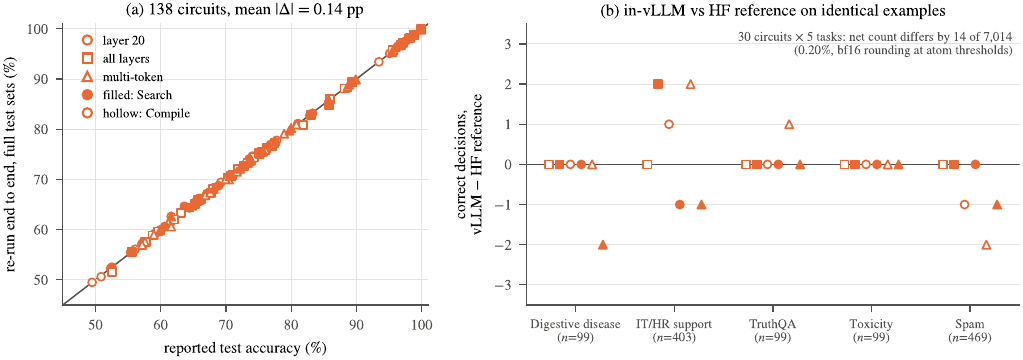}
\caption{\textbf{Serving preserves accuracy.} (a)~Five-gate circuits re-run end
         to end from raw text on the full test sets (Hugging Face reference)
         against the reported accuracies; shape: access level, fill: technique.
         (b)~Correct decisions inside vLLM minus the reference on identical
         examples, per circuit.}
\label{fig:serving-accuracy}
\end{figure*}

\para{Verification}
We verify \bitprobeserve{} in two steps.
First, for each \bitprobe{} we feed identical binary
features to its truth table and a PyTorch implementation
of its rule, and confirm matching outputs.
Second, we compare end-to-end predictions against a
Hugging Face reference.
The reference reproduces the reported test accuracies
within 0.14 percentage points on average across
138 \bitprobes{} and 23 tasks.
Inside vLLM, 14 of 7{,}014 predictions (0.20\%) differ
from the reference, consistent with BF16 rounding near
feature thresholds (\cref{fig:serving-accuracy}).
GPU execution is not always bit-exact
\citep{badash2026hawkeye}.
\section{Full Per-Task Results}
\label{app:per-task}

\begin{figure*}[!htbp]
\centering
\includegraphics[width=\textwidth]
{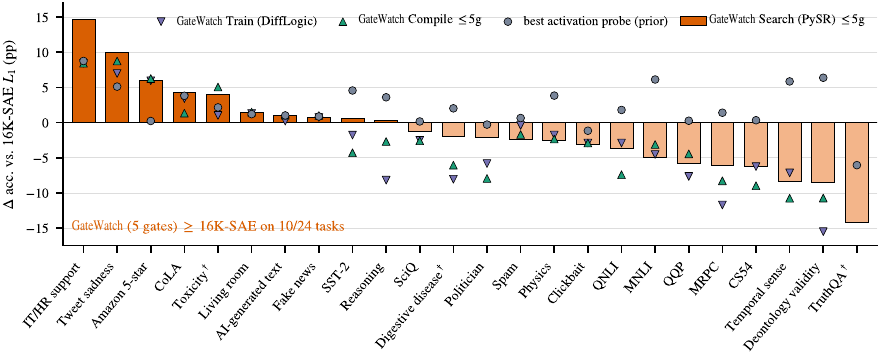}
\caption{\textbf{Per-task accuracy difference to the 16K-SAE $L_1$ probe},
         sorted (validation-selected). Bars: five-gate Search \bitprobe{}s (dark:
         at least as accurate as the SAE probe). Markers: Train and five-gate
         Compile \bitprobe{}s, and the best prior activation head. \bitprobe{}s win on
         content and affect tasks (left) and lose on tasks that compare two
         text segments or require multi-step reasoning (right;
         \cref{sec:ablations}). $^{\dagger}$99 test examples.}
\label{fig:pertask}
\end{figure*}

\begin{table}[!htbp]
\centering
\caption{\textbf{24-task summary} (validation-selected). Mean test accuracy,
         memory per probe, and win/tie/loss (W/T/L) against the 16K-SAE $L_1$
         probe. Prior work (top) uses layer-20 last-token inputs; \bitprobe{}s
         (bottom) use multi-token binary features from all layers. Shaded rows
         are $\le$5-gate \bitprobe{}s.}
\label{tab:aggregate-results-full}
\footnotesize\setlength{\tabcolsep}{1.4pt}
\begin{tabular}{@{}lrrrrrc@{}}
\toprule
Method & Gates & Live & KiB & Acc.\,\% & $\Delta$\,pp & W/T/L \\
\midrule
\grp{7}{Prior work (SAE-Probes, layer 20, last token)} \\
Logistic reg. & -- & -- & 112.6 & 83.76 & $+$2.10 & 21/0/3 \\
PCA + LR & -- & -- & 1{,}495.0 & 82.04 & $+$0.38 & 18/0/6 \\
$k$-NN & -- & -- & 14{,}428.0 & 75.76 & $-$5.90 & 3/0/21 \\
XGBoost & -- & -- & 314.3 & 82.11 & $+$0.45 & 16/0/8 \\
MLP (3-layer) & -- & -- & 254.0 & 83.82 & $+$2.16 & 21/1/2 \\
Best act.\ head$^{\ast}$ & -- & -- & 987.4 & 83.84 & $+$2.19 & 21/0/3 \\
\dashrule
16K-SAE $L_1$ & -- & 16{,}384 & 115{,}261.4 & 81.66 & \phantom{$+$}0.00 & -- \\
\midrule
\grp{7}{\bitprobe{} (ours)} \\
Integer LR (L20) & 6{,}157$^{\dagger}$ & $\le$256 & 1{,}009.7 & 79.17 & $-$2.49 &
4/0/20 \\
Exact circ.\ (L20) & 6{,}157$^{\dagger}$ & $\le$256 & 1{,}033.5 & 79.17 & $-$2.49 &
4/0/20 \\
\dashrule
Train (DiffLogic) & 2.2$^{\ddagger}$ & 3.1 & 191.5 & 78.42 & $-$3.23 & 8/0/16 \\
Compile $\leq$10g & 5.6 & 5.3 & 36.8 & 78.17 & $-$3.49 & 8/0/16 \\
Search $\leq$10g & 5.5 & 5.4 & 38.0 & 80.77 & $-$0.89 & 10/0/14 \\
\rowcolor{ours}Compile $\leq$5g & 3.5 & 4.0 & 28.0 & 78.48 & $-$3.18 & 8/0/16
\\
\rowcolor{ours}Search $\leq$5g & 3.5 & 3.8 & \textbf{26.3} & \textbf{80.49} &
\textbf{$-$1.17} & 10/0/14 \\
\bottomrule
\end{tabular}
\\[2pt]
{\footnotesize\raggedright $^{\ast}$Best of the five heads per task, chosen on
validation. $^{\dagger}$Mean exact-circuit size (median 5{,}513; range
9--12{,}383). $^{\ddagger}$DiffLogic gates are 16-way two-input gates; its
memory conservatively keeps all $K$ selected rows (about 21~KiB with only its live
rows). ``Live'': distinct SAE rows that reach the output.\par}
\end{table}

\begin{table*}[!htbp]
\centering
\caption{Test accuracy (\%) on all 24 tasks, validation-selected (the full
         version of \cref{tab:aggregate-results-full}). Left of the dashed rule: prior
         work (layer 20, last token). Right: \bitprobe{}, multi-token binary features
         from all layers. \colorbox{win}{Green}: $\ge$ 16K-SAE probe;
         \textbf{bold}: best in row. $^{\dagger}$99 test examples. $^{\S}$Test
         split is 98.8\% one class.}
\label{tab:per-task-full}
\small\setlength{\tabcolsep}{4.2pt}
\begin{tabular}{@{}lr|rrr;{2pt/2pt}rrrrr@{}}
\toprule
& & \multicolumn{3}{c;{2pt/2pt}}{\textit{Prior work}} &
\multicolumn{5}{c}{\textit{\bitprobe{} (ours), all layers, multi-token}} \\
\cmidrule(lr){3-5}\cmidrule(l){6-10}
Task & $n_{\text{test}}$ & Best act. & MLP & 16K-SAE & Train & Compile $\le$10g
& Search $\le$10g & Compile $\le$5g & Search $\le$5g \\
\midrule
CS54 & 3{,}975 & 66.47 & \textbf{67.57} & 66.14 & 59.87 & 57.64 & 61.43 & 57.16
& 59.90 \\
Living room & 2{,}093 & 99.62 & 99.24 & 98.42 & \cellcolor{win}99.71 &
\cellcolor{win}\textbf{99.95} & \cellcolor{win}99.81 & \cellcolor{win}99.90 &
\cellcolor{win}99.81 \\
SciQ & 3{,}975 & 91.37 & \textbf{92.48} & 91.22 & 88.68 & 88.28 & 89.28 & 88.63
& 89.91 \\
QNLI & 3{,}975 & \textbf{85.26} & 84.63 & 83.47 & 80.55 & 81.21 & 81.79 & 76.05
& 79.72 \\
Toxicity$^{\dagger}$ & 99 & 87.00 & 87.88 & 84.85 & \cellcolor{win}85.86 &
\cellcolor{win}\textbf{89.90} & \cellcolor{win}88.89 &
\cellcolor{win}\textbf{89.90} & \cellcolor{win}88.89 \\
Tweet sadness & 1{,}475 & 63.21 & 64.07 & 58.10 & \cellcolor{win}65.08 &
\cellcolor{win}67.46 & \cellcolor{win}66.78 & \cellcolor{win}66.85 &
\cellcolor{win}\textbf{68.14} \\
MNLI & 2{,}307 & \textbf{84.88} & 83.66 & 78.76 & 74.21 & 76.29 & 73.86 & 75.64
& 73.86 \\
Politician & 3{,}975 & 66.62 & 66.74 & \textbf{66.92} & 61.08 & 62.89 & 64.83 &
58.97 & 64.83 \\
CoLA & 3{,}975 & 79.48 & 76.88 & 75.67 & \cellcolor{win}79.07 &
\cellcolor{win}77.11 & \cellcolor{win}\textbf{80.00} & \cellcolor{win}77.01 &
\cellcolor{win}\textbf{80.00} \\
TruthQA$^{\dagger}$ & 99 & 80.81 & \textbf{87.88} & 86.87 & 68.69 & 61.62 &
66.67 & 61.62 & 72.73 \\
Temporal sense & 1{,}199 & \textbf{89.92} & 89.74 & 84.07 & 76.90 & 74.06 &
79.82 & 73.31 & 75.73 \\
Physics & 3{,}975 & \textbf{79.85} & 78.89 & 76.03 & 74.24 & 74.06 & 74.16 &
73.69 & 73.48 \\
Reasoning & 1{,}671 & \textbf{73.91} & 73.01 & 70.32 & 62.12 & 68.40 &
\cellcolor{win}71.27 & 67.62 & \cellcolor{win}70.56 \\
Deontology validity & 3{,}975 & \textbf{87.07} & 87.07 & 80.70 & 65.18 & 67.42 &
74.39 & 69.96 & 72.18 \\
MRPC & 1{,}363 & 81.36 & \textbf{82.47} & 79.97 & 68.23 & 71.68 & 73.07 & 71.68
& 73.88 \\
QQP & 3{,}975 & 83.63 & \textbf{83.97} & 83.37 & 75.70 & 68.60 & 79.42 & 78.92 &
77.56 \\
SST-2 & 3{,}975 & \textbf{89.66} & 88.20 & 85.11 & 83.32 & 80.81 &
\cellcolor{win}86.92 & 80.81 & \cellcolor{win}85.76 \\
AI-generated text$^{\S}$ & 3{,}975 & 98.82 & 98.29 & 97.81 &
\cellcolor{win}98.01 & \cellcolor{win}\textbf{98.84} &
\cellcolor{win}\textbf{98.84} & \cellcolor{win}\textbf{98.84} &
\cellcolor{win}\textbf{98.84} \\
Spam & 469 & \textbf{98.94} & 98.29 & 98.29 & 97.87 & 96.38 & 95.95 & 96.59 &
95.95 \\
Fake news & 2{,}095 & 99.86 & 99.95 & 99.00 & \cellcolor{win}99.71 &
\cellcolor{win}\textbf{100.00} & \cellcolor{win}99.71 &
\cellcolor{win}\textbf{100.00} & \cellcolor{win}99.71 \\
Clickbait & 1{,}543 & 98.44 & \textbf{99.74} & 99.61 & 96.69 & 96.69 & 96.57 &
96.69 & 96.57 \\
Digestive disease$^{\dagger}$ & 99 & \textbf{65.66} & \textbf{65.66} & 63.64 &
55.56 & 56.57 & 61.62 & 57.58 & 61.62 \\
IT/HR support & 403 & 70.79 & 66.50 & 62.03 & \cellcolor{win}70.47 &
\cellcolor{win}64.52 & \cellcolor{win}\textbf{77.17} & \cellcolor{win}70.47 &
\cellcolor{win}76.67 \\
Amazon 5-star & 975 & 89.65 & 88.82 & 89.44 & \cellcolor{win}95.38 &
\cellcolor{win}95.69 & \cellcolor{win}\textbf{96.31} & \cellcolor{win}95.69 &
\cellcolor{win}95.49 \\
\midrule
Mean (all 24) & -- & \textbf{83.84} & 83.82 & 81.66 & 78.42 & 78.17 & 80.77 &
78.48 & 80.49 \\
Median (all 24) & -- & 85.07 & 85.85 & 83.42 & 76.30 & 75.18 & 79.62 & 75.84 &
77.12 \\
W/T/L vs.\ SAE & -- & 21/0/3 & 21/1/2 & -- & 8/0/16 & 8/0/16 & 10/0/14 & 8/0/16
& 10/0/14 \\
\dashrule
Mean MiB & -- & 0.964 & 0.248 & 112.560 & 0.187 & 0.036 & 0.037 & 0.027 & 0.026
\\
\bottomrule
\end{tabular}
\end{table*}
\
\Cref{tab:app-ablation} gives the per-task numbers behind\
\cref{tab:ablation-access}.\
The illustrative circuits in \cref{fig:circuits} include\
examples from the all-layer, last-token configuration\
reported in this table. Their accuracies may therefore\
differ from the multi-token results in \cref{tab:per-task-full}.\
\
\begin{table*}[!htbp]
\centering
\caption{Test accuracy (\%) for the L20 and all-layer (last-token) settings,
         validation-selected. ``Gates'' is the size of the exact L20 circuit.}
\label{tab:app-ablation}
\scriptsize\setlength{\tabcolsep}{3pt}
\begin{tabular}{@{}lr rr;{2pt/2pt}rrrrr;{2pt/2pt}rrrrr@{}}
\toprule
& & \multicolumn{7}{c;{2pt/2pt}}{\textit{Layer 20, last token}} &
\multicolumn{5}{c}{\textit{All layers, last token}} \\
\cmidrule(lr){3-9}\cmidrule(l){10-14}
Task & Gates & Int.\ LR & Exact & Train & Comp.$\le$10 & Search$\le$10 &
Comp.$\le$5 & Search$\le$5 & Train & Comp.$\le$10 & Search$\le$10 & Comp.$\le$5
& Search$\le$5 \\
\midrule
CS54 & 10{,}663 & 62.21 & 62.21 & 59.72 & 60.38 & 55.40 & 59.60 & 55.40 & 54.06
& 57.76 & 63.27 & 57.76 & 64.43 \\
Living room & 144 & 97.18 & 97.18 & 96.61 & 97.13 & 97.13 & 97.13 & 97.13 &
99.71 & 98.09 & 99.90 & 98.09 & 99.90 \\
SciQ & 5{,}056 & 90.21 & 90.21 & 82.92 & 81.51 & 85.81 & 81.06 & 81.01 & 86.64 &
87.52 & 89.33 & 85.91 & 89.33 \\
QNLI & 9{,}208 & 80.10 & 80.10 & 76.96 & 74.77 & 73.41 & 73.71 & 73.41 & 77.74 &
76.98 & 80.83 & 75.55 & 75.14 \\
Toxicity & 5{,}087 & 87.88 & 87.88 & 81.82 & 75.76 & 81.82 & 77.78 & 72.73 &
84.85 & 81.82 & 87.88 & 81.82 & 85.86 \\
Tweet sadness & 11{,}891 & 59.93 & 59.93 & 61.63 & 50.03 & 58.37 & 52.34 & 57.36
& 65.69 & 59.19 & 65.90 & 58.78 & 65.90 \\
MNLI & 4{,}643 & 74.17 & 74.17 & 72.35 & 59.30 & 70.78 & 56.09 & 68.88 & 68.57 &
74.82 & 75.12 & 73.34 & 75.12 \\
Politician & 817 & 66.82 & 66.82 & 39.92 & 58.64 & 64.38 & 57.11 & 64.38 & 60.75
& 62.47 & 65.69 & 62.01 & 65.69 \\
CoLA & 5{,}540 & 73.33 & 73.33 & 69.81 & 70.67 & 74.14 & 70.69 & 74.14 & 75.14 &
69.31 & 76.23 & 74.31 & 76.23 \\
TruthQA & 5{,}921 & 80.81 & 80.81 & 68.69 & 64.65 & 70.71 & 61.62 & 63.64 &
72.73 & 65.66 & 71.72 & 65.66 & 72.73 \\
Temporal sense & 10{,}808 & 86.24 & 86.24 & 72.81 & 74.48 & 73.56 & 74.06 &
73.56 & 77.81 & 76.56 & 78.57 & 71.89 & 77.15 \\
Physics & 1{,}323 & 70.31 & 70.31 & 68.48 & 57.84 & 66.01 & 59.97 & 66.01 &
72.25 & 69.61 & 73.53 & 70.87 & 70.99 \\
Reasoning & 12{,}383 & 66.91 & 66.91 & 51.35 & 58.83 & 65.47 & 57.39 & 65.47 &
66.61 & 63.79 & 68.28 & 63.14 & 65.23 \\
Deontology validity & 10{,}902 & 78.19 & 78.19 & 57.21 & 58.64 & 69.36 & 59.12 &
60.63 & 67.25 & 66.21 & 72.43 & 66.21 & 70.24 \\
MRPC & 9{,}968 & 74.98 & 74.98 & 60.90 & 62.00 & 69.11 & 61.48 & 68.16 & 69.26 &
61.48 & 70.36 & 59.65 & 70.80 \\
QQP & 5{,}930 & 77.94 & 77.94 & 77.99 & 68.63 & 77.99 & 69.23 & 76.81 & 75.70 &
67.92 & 72.75 & 68.03 & 70.99 \\
SST-2 & 5{,}420 & 78.04 & 78.04 & 70.97 & 70.87 & 75.32 & 67.07 & 73.11 & 81.21
& 76.86 & 83.50 & 77.43 & 83.07 \\
AI-generated text & 1{,}692 & 97.18 & 97.18 & 96.13 & 95.32 & 96.78 & 93.46 &
96.78 & 98.36 & 98.84 & 98.84 & 98.84 & 98.84 \\
Spam & 1{,}235 & 96.59 & 96.59 & 95.52 & 94.24 & 95.52 & 95.10 & 95.52 & 94.03 &
95.52 & 96.80 & 95.52 & 96.80 \\
Fake news & 9 & 98.90 & 98.90 & 91.46 & 98.90 & 99.24 & 98.90 & 99.24 & 99.71 &
100.00 & 99.71 & 100.00 & 99.71 \\
Clickbait & 1{,}744 & 98.38 & 98.38 & 95.98 & 95.92 & 97.21 & 96.18 & 97.60 &
95.85 & 97.15 & 98.25 & 97.15 & 98.25 \\
Digestive disease & 10{,}776 & 53.54 & 53.54 & 53.54 & 49.49 & 47.47 & 49.49 &
52.53 & 48.48 & 54.55 & 54.55 & 52.53 & 55.56 \\
IT/HR support & 11{,}116 & 66.25 & 66.25 & 58.56 & 51.12 & 60.05 & 50.87 & 60.05
& 72.70 & 58.31 & 68.73 & 57.57 & 67.74 \\
Amazon 5-star & 5{,}485 & 83.90 & 83.90 & 78.97 & 83.79 & 82.77 & 83.28 & 82.87
& 87.69 & 86.77 & 90.05 & 88.21 & 89.33 \\
\midrule
Mean & 6{,}157 & 79.17 & 79.17 & 72.51 & 71.37 & 75.33 & 70.95 & 74.02 & 77.20 &
75.30 & 79.26 & 75.01 & 78.54 \\
\bottomrule
\end{tabular}
\end{table*}

\FloatBarrier

\end{document}